\documentclass[english,man,floatsintext]{apa6}
\usepackage[LGR,T1]{fontenc}
\usepackage[latin9]{inputenc}
\usepackage{array}
\usepackage{verbatim}
\usepackage{rotating}
\usepackage{booktabs}
\usepackage{textcomp}
\usepackage{url}
\usepackage{multirow}
\usepackage{amstext}
\usepackage{amssymb}
\usepackage{stmaryrd}
\usepackage{stackrel}
\usepackage{graphicx}
\usepackage{setspace}
\usepackage[authoryear]{natbib}
\makeatletter

\DeclareRobustCommand{\greektext}{%
  \fontencoding{LGR}\selectfont\def\encodingdefault{LGR}}
\DeclareRobustCommand{\textgreek}[1]{\leavevmode{\greektext #1}}
\ProvideTextCommand{\~}{LGR}[1]{\char126#1}

\providecommand{\tabularnewline}{\\}

\@ifundefined{date}{}{\date{}}
\@ifundefined{showcaptionsetup}{}{%
 \PassOptionsToPackage{caption=false}{subfig}}
\usepackage{subfig}
\makeatother

\usepackage{babel}
\begin{document}
\title{Modeling citation worthiness by using attention-based Bidirectional
Long Short-Term Memory networks and interpretable models}
\shorttitle{Modeling citation worthiness}
\author{Tong Zeng$^{1,2}$ and Daniel E. Acuna$^{2,*}$}
\note{$^{*}$Corresponding author: \url{deacuna@syr.edu}}
\affiliation{$^{1}$School of Information Management, Nanjing University, Nanjing,
China\\
$^{2}$School of Information Studies, Syracuse University, Syracuse,
New York, USA}
\abstract{Scientist learn early on how to cite scientific sources to support
their claims. Sometimes, however, scientists have challenges determining
where a citation should be situated---or, even worse, fail to cite
a source altogether. Automatically detecting sentences that need a
citation (i.e., \emph{citation worthiness}) could solve both of these
issues, leading to more robust and well-constructed scientific arguments.
Previous researchers have applied machine learning to this task but
have used small datasets and models that do not take advantage of
recent algorithmic developments such as attention mechanisms in deep
learning. We hypothesize that we can develop significantly accurate
deep learning architectures that learn from large supervised datasets
constructed from open access publications. In this work, we propose
a Bidirectional Long Short-Term Memory (BiLSTM) network with attention
mechanism and contextual information to detect sentences that need
citations. We also produce a new, large dataset (PMOA-CITE) based
on PubMed Open Access Subset, which is orders of magnitude larger
than previous datasets. Our experiments show that our architecture
achieves state of the art performance on the standard ACL-ARC dataset
($F_{1}=0.507$) and exhibits high performance ($F_{1}=0.856$) on
the new PMOA-CITE. Moreover, we show that it can transfer learning
across these datasets. We further use interpretable models to illuminate
how specific language is used to promote and inhibit citations. We
discover that sections and surrounding sentences are crucial for our
improved predictions. We further examined purported mispredictions
of the model, and uncovered systematic human mistakes in citation
behavior and source data. This opens the door for our model to check
documents during pre-submission and pre-archival procedures. We discuss
limitations of our work and make this new dataset, the code, and a
web-based tool available to the community.}
\maketitle

\section{Introduction}

Scientists and journalists have challenges determining proper citations
in the ever increasing sea of information. More fundamentally, when
and where a citation is needed---sometimes called \emph{citation
worthiness---}is a crucial first step to solve this challenge. In
the general media, some problematic stories have shown that claims
need citations to make them verifiable---e.g., \textit{\emph{the
debunked }}\textit{A Rape on Campus} article in the \textit{Rolling
Stone} magazine \citep{wiki:2018}. Analyses of Wikipedia have revealed
that lack of citations correlates with an article's immaturity \citep{jack2014citation,chen2012citation}.
In science, the lack of citations leaves readers wondering how results
were built upon previous work \citep{aksnes2009researchers}. Also,
it precludes researchers from getting appropriate credit, important
during hiring and promotion \citep{gazni2016author}. The sentences
surrounding a citation\emph{ }provide rich information for common
semantic analyses, such as information retrieval \citep{nakov2004citances}.
There should be methods and tools to help scientists cite; in this
work, we want to understand where citations should be situated in
a paper with the goal of automatically suggesting them.

We first review a closely related problem: citation recommendation.
Several research groups have studied how to recommend citations at
the article and local levels, separately. At the article level, \citet{kuccuktuncc2012direction}
uses graph based methods for estimating citation relationships between
papers. \citet{mcnee2002recommending} and \citet{torres2004enhancing}
use collaborative filtering to make such suggestions by bootstrapping
on collective citation patterns. These techniques work well for making
article-level citation recommendations and they frequently rely on
knowing where a citation should be located. At the local level, \citet{HeContextawareCitationRecommendation2010}
propose context aware citation recommendation by using local contextual
information of the places where citations are made. More recently,
other groups have used more sophisticated neural network models to
estimate a semantic representation of sentences---e.g., \citet{HuangNeuralProbabilisticModela}
use distributed representations, \citet{ebesu2017neural} use auto-encoders,
and \citet{BhagavatulaContentBasedCitationRecommendation2018} use
a two-step process to first embed documents into a vector representation
and then rank them according to a relevance estimation task. These
techniques have shown that it is possible to provide detailed sentence
level suggestions if the place to put a citation is already known.
This implies that detecting which sentences need a citation is a crucial
first step for sentence-level citation recommendation.

Relatively much less work has been done on detecting where a citation
should be. \citet{he2011citation} were the first to introduce the
task of identifying candidate location where citations are needed
in the context of scientific articles. \citet{jack2014citation} studied
how to detect citation needs in Wikipedia. \citet{peng2016news} used
the learning-to-rank framework to solve citation recommendation in
news articles. These are very diverse domains, and therefore it is
difficult to generalize results. We contend that a large standard
dataset of citation location with open code and services would significantly
improve the systematic study of the problem. Thus, the task of citation
worthiness detection is relatively new and needs further exploration.

Recently, Recurrent Neural Networks (RNN) and Convolutional Neural
Networks (CNN) architectures have been used for the detection of citation
worthiness. The group of \citet{bonabCitationWorthinessSentences2018}
applied CNN based classifiers, achieving state of art performance.
\citet{farber2018cite} proposed stacking a RNN on top of a CNN, achieving
good performance as well. However, one of the problems with RNN networks
is that they tend to dismiss long-distance dependencies. The attention
mechanism has been shown to fix some of this issue, and it can potentially
help for the detection of citation worthiness.

The attention mechanism is a relatively recent development in neural
networks motivated by human visual attention. Humans get more information
from the region they pay attention to, and perceive less from other
regions. An attention mechanism in neural networks was first introduced
in computer vision \citep{sunObjectbasedVisualAttention2003}, and
later applied to NLP for machine translation \citep{bahdanauNeuralMachineTranslation2014}.
Attention has quickly become adopted in other sub-domains. \citet{luongEffectiveApproachesAttentionbased2015}
examined several attention scoring functions for machine translation.
\citet{liDatasetNeuralRecurrent2016} used attention mechanisms to
improve results in a question-answering task. \citet{zhouAttentionBasedBidirectionalLong2016}
made use of an attention-based LSTM network to do relational classification.
\citet{linStructuredSelfattentiveSentence2017} used attention to
improve sentence embedding. Recently, \citet{vaswaniAttentionAllYou2017a}
built an architecture called \emph{transformer} that promises to replace
recurrent neural networks (RNNs) altogether by only using attention
mechanisms. These results show the advantage of attention for NLP
tasks and thus its potential benefit for citation worthiness.

In this study, we formulate the detection of sentences that need citations
as a classification task that can be effectively solved with a deep
learning architecture that relies on an attention mechanism. Our contributions
are the following:
\begin{APAenumerate}
\item A deep learning architecture based on bidirectional LSTM with attention
and contextual information for citation worthiness
\item A new large scale dataset for the citation worthiness task that is
300 times bigger that the next current alternative
\item A set of classic interpretable models that provide insights into the
language used for making citations
\item An examination of common citation mistakes---from unintentional omissions
to potentially problematic mis-citations
\item An evaluation of transfer learning between our proposed dataset and
the ACL-ARC dataset
\item The code to produce the dataset and results, a web-based tool for
the community to evaluate our predictions, and the pre-processed dataset.
\end{APAenumerate}

\section{Materials and methods}

\subsection{Problem Formulation}

We consider a paper as a sequence of sentences, and sentences as a
sequence of words. We denote a paper as $D$, a word as $w$, and
a sentence as $S$, $S$$=[w_{1},w_{2},...,w_{N}]$. The \textit{citation
location} is the location where the paper cites a reference within
the sequence of words---e.g., \textquotedbl{[}6,8{]}\textquotedbl .
The \textit{citing sentence} is a sentence that contains one or more
citations, and we denote it as $S^{c}$. A non-citing sentence is
denoted as $S^{nc}$. Finally, \emph{citation context} could be any
information describing the context of a sentence. These definitions
will be used throughout this article.

There are multiple ways of defining a citation context. A frequently
employed approach is to define the citation context as a sequence
of words around citation locations \citep{HuangNeuralProbabilisticModela,mikolov2013distributed,duma2016applying}.
The length of such sequence may vary from paper to paper---\citet{HeContextawareCitationRecommendation2010}
specified the context as a fixed window of 100 words; \citet{duma2014citation}
experimented with 5, 10, 20 and 30 words. However, the number of words
associated with a citation may differ on a case-by-case basis \citep{ritchie2009citation},
and arbitrarily truncating a sentence due to the size of a window
could reduce the strength of contextual signal. As others have observed
\citep[i.e.,][]{allerton1969sentence,frajzyngier2005linguistic,halliday2014introduction},
humans use a sentence as the fundamental unit to express thoughts.
Therefore, we will use sentences as the minimum unit for our algorithm.
While some researchers only consider the citing sentence as the context
\citep{he2012position}, we also consider the previous and next sentences
as citation context. Furthermore, we observe that the section $Sec$
(e.g. introduction, methods, results, discussion, and so forth) may
affect whether a sentence needs a citation. Therefore, we include
section as part of the context. The context, then, is denoted by $CC=\{S_{n-1},S_{n}^{c},S_{n+1},Sec\}$.

\begin{figure}[ht]
\begin{centering}
\includegraphics[width=0.8\linewidth]{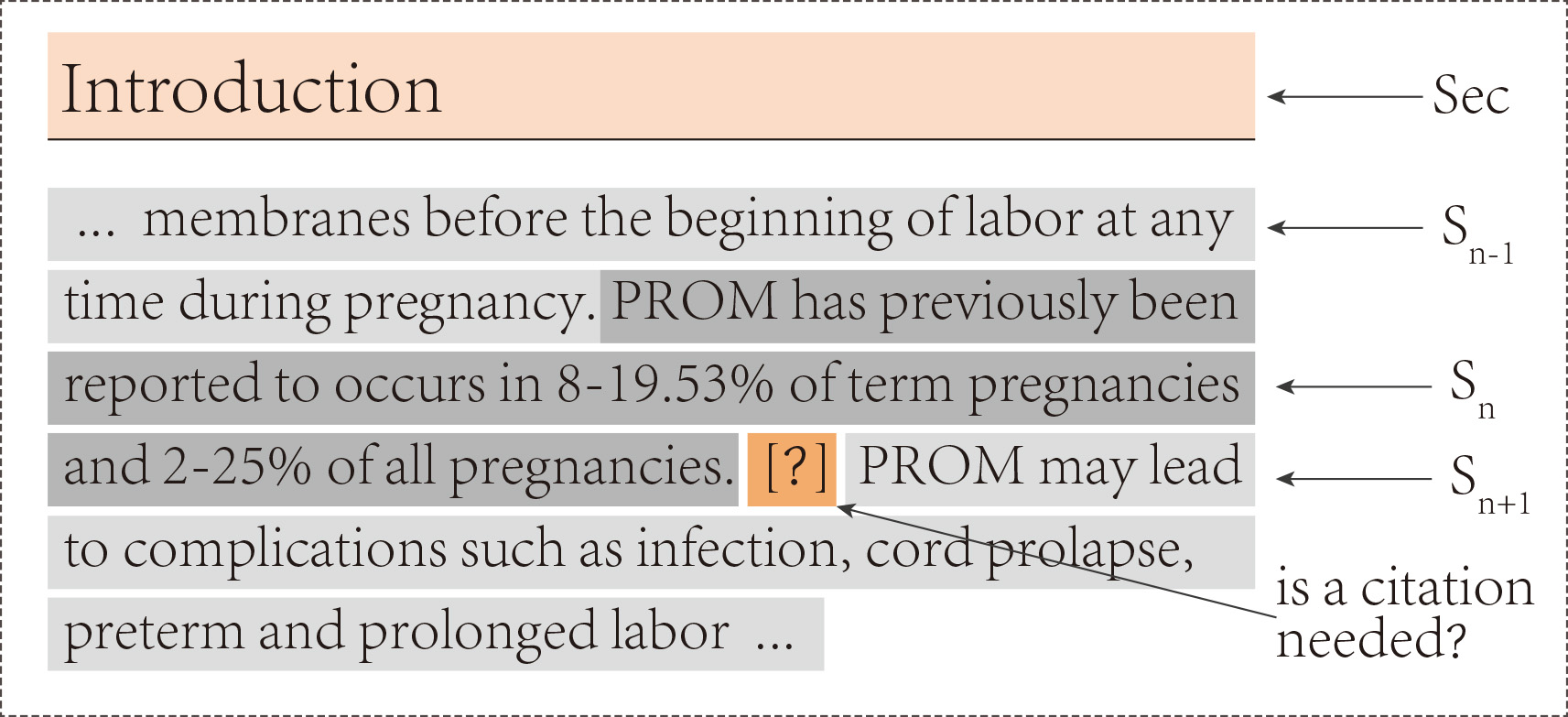}
\par\end{centering}
\caption{\label{fig:citation_needed} Citation worthiness prediction problem.
For a given sentence ($S_{n}$), the goal of the task is to predict
whether it needs a citation. The prediction task may use the section,
the previous and next sentences (i.e., $S_{n-1}$ and $S_{n+1}$)
for such prediction.}
\end{figure}

We now state the citation worthiness problem. The user submits a manuscript
without reference list and without citation placeholders. Our goal
is to predict for each sentence whether it needs a citation (Fig.
\ref{fig:citation_needed}). Since there are only two outputs, we
cast this prediction as a binary classification task.

\subsection{Data sources and data pre-processing}

\subsubsection{ACL Anthology Reference Corpus}

The ACL Anthology Reference Corpus (ACL-ARC) is a collection of scientific
articles in Computational Linguistics. The ACL-ARC 1.0 dataset consists
of 10,921 articles up to February 2007, including the source PDF,
automatically extracted full text, and the metadata for the articles.
In order to use the ACL-ARC dataset, we need to remove some noisy
sentences, such as footnotes, mathematical equations, and URLs. \citet{bonabCitationWorthinessSentences2018}
carried out all these pre-processing steps and made the data available
on the Internet\footnote{\url{https://ciir.cs.umass.edu/downloads/sigir18_citation/}}.
This dataset consists of 85,778 sentences with citations and 1,142,275
sentences without citations. More statistics are presented in Table
\ref{table:characteristics-pmoa}.

\subsubsection{PubMed Central Open Access Subset}

\begin{figure}[ht]
\begin{centering}
\includegraphics[width=0.8\linewidth]{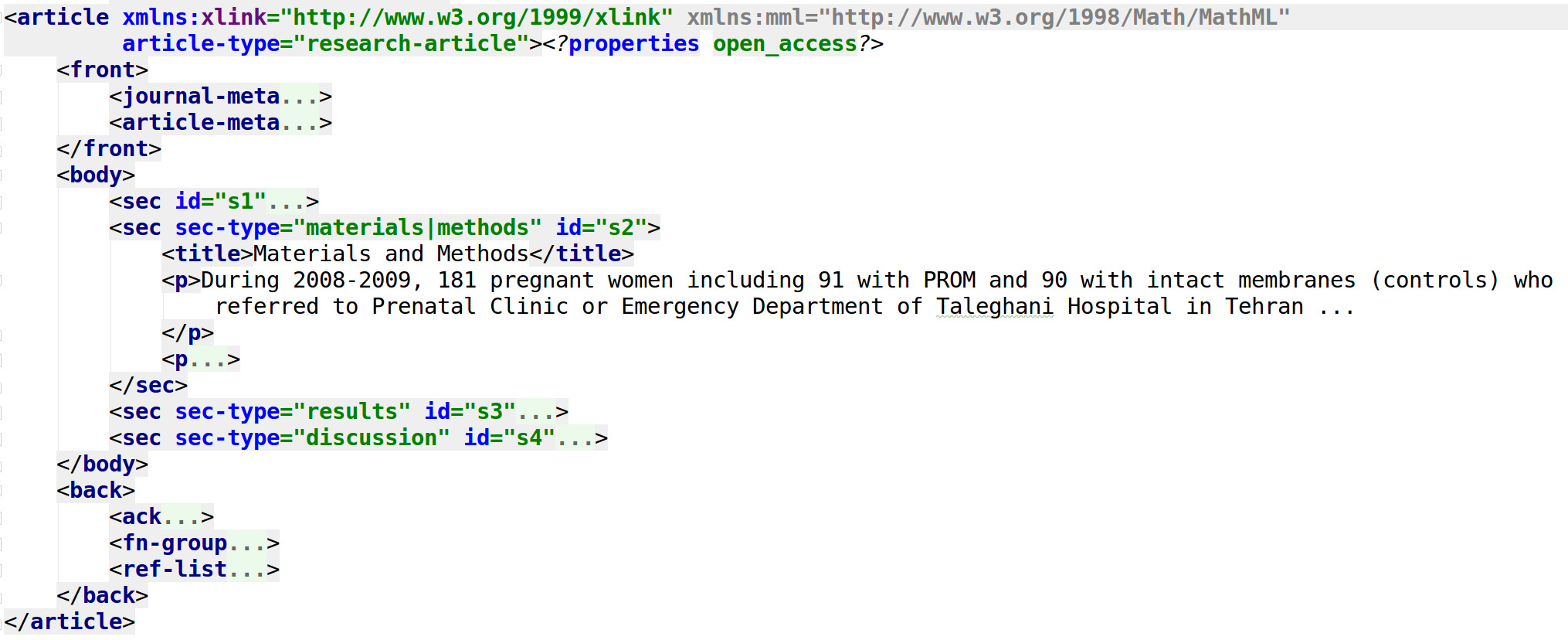}
\par\end{centering}
\caption{\label{fig:sample-pmc}A sample of a PMC Open Access Subset (PMOAS)
XML. The structure is defined by a standard Document Type Definition
(DTD) which makes all articles consistent. In particular, the tag
and attributes of a citation are well known.}
\end{figure}

PubMed Central Open Access subset (PMOAS) is a full-text collection
of scientific literature in bio-medical and life sciences. PMOAS is
created by the US's National Institutes of Health. We obtain a snapshot
of PMOAS on August, 2019. The dataset consists of more than 2 million
full-text journal articles organized in well-structured XML files
(Fig. \ref{fig:sample-pmc}). The XML format follow the Journal Article
Tag Suite (JATS) which developed by the National Information Standards
Organization \citep{JATS2013}.

\begin{figure}
\begin{centering}
\subfloat[Character count distribution of sentences]{\begin{centering}
\includegraphics[width=0.4\linewidth]{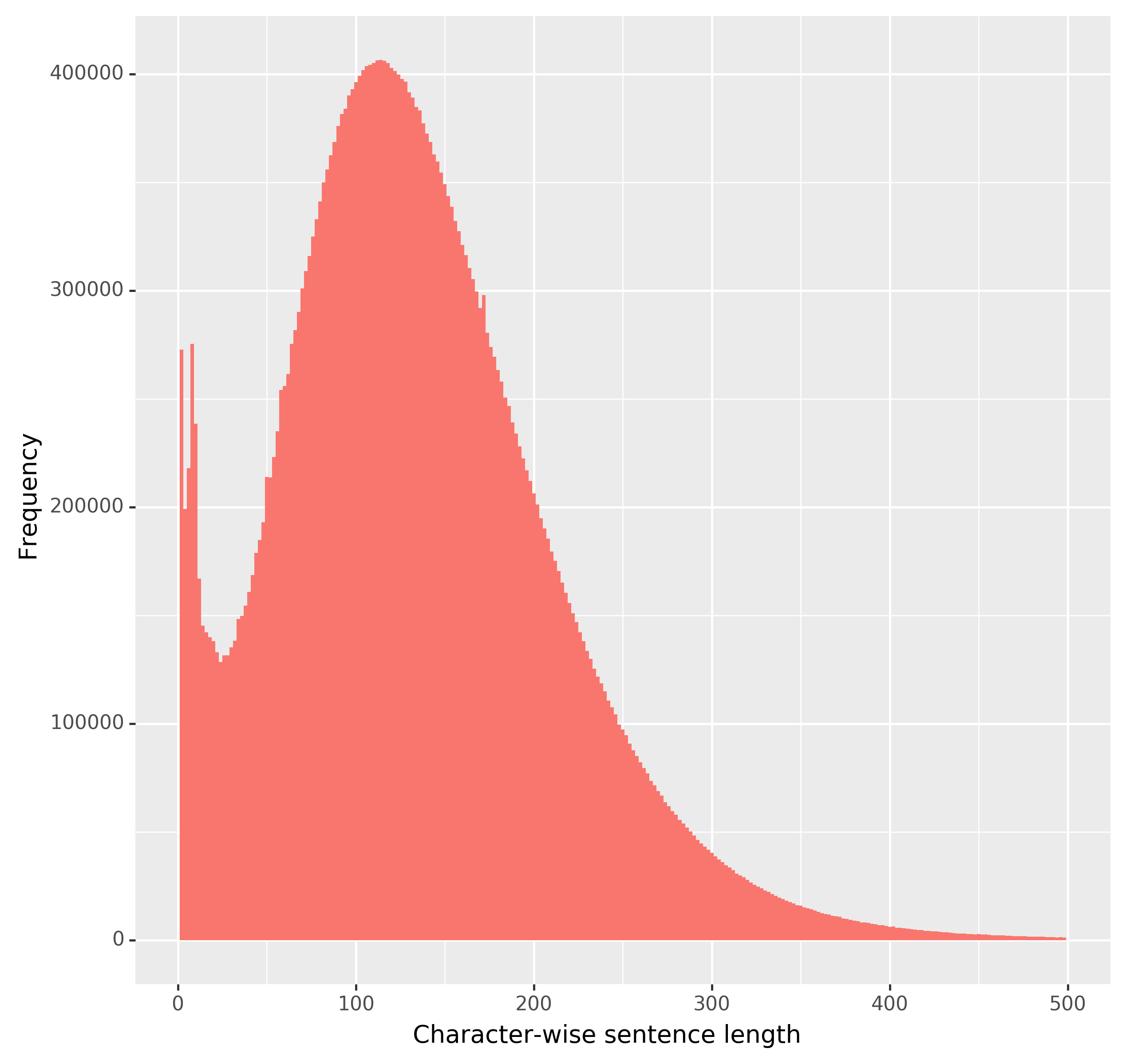}
\par\end{centering}
}\subfloat[Word count distribution of sentences]{
\centering{}\includegraphics[width=0.4\linewidth]{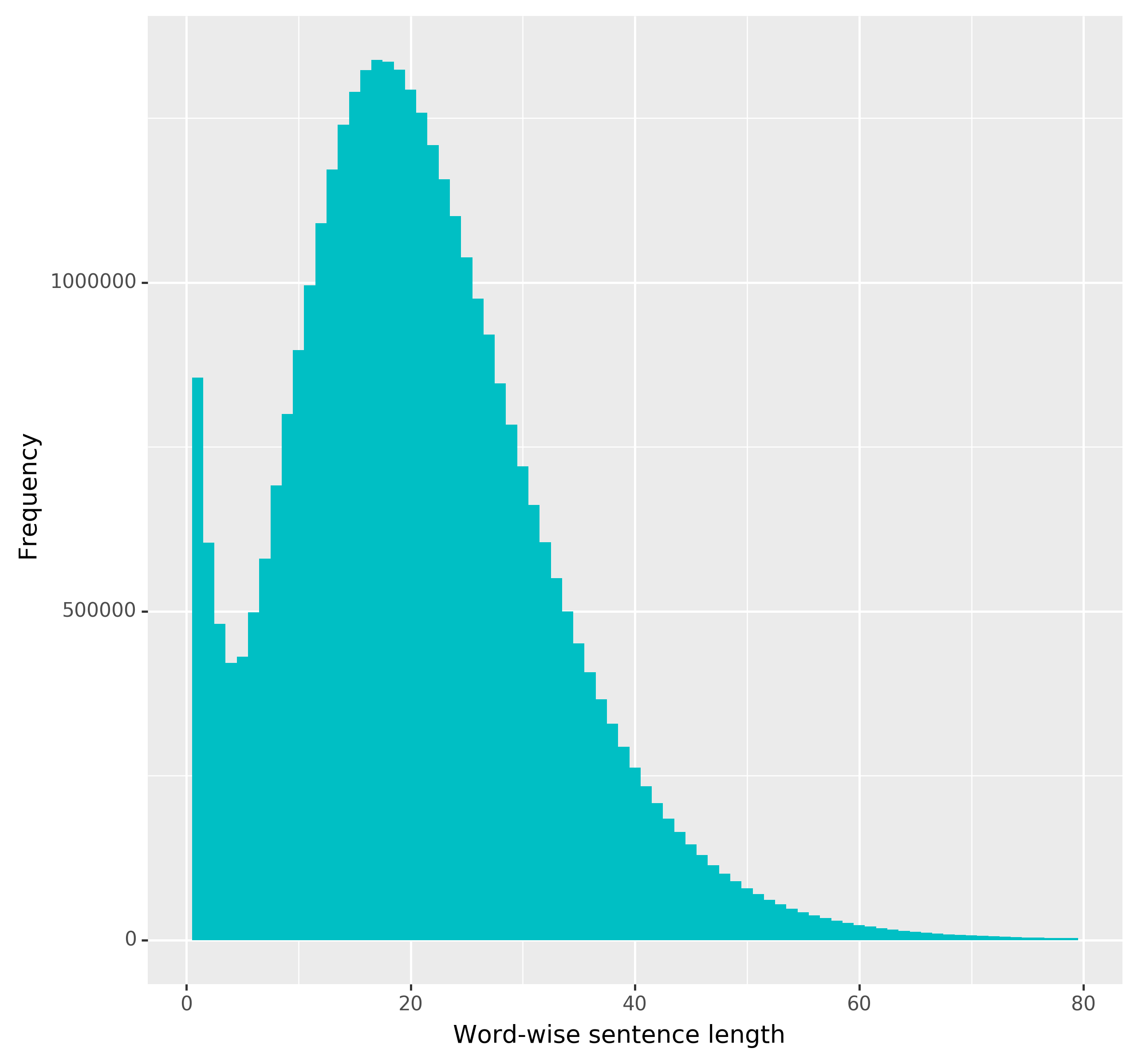}}
\par\end{centering}
\caption{\label{fig:The-distribution-of-sent-len}The distribution of sentence
length}

\end{figure}

We now describe how we prepare the dataset.
\begin{APAenumerate}
\item \textbf{Sentence segmentation and outlier removal}. Text in a PMOAS
XML file is marked by a paragraph tag, but there might be other XML
tags inside paragraph tags. Therefore, we needed to get all the text
of a paragraph from XML tags recursively and break paragraphs into
sentences. We used spaCy Python package to do the sentence splitting
\citep{spacy2}. However, there are some outliers in the sentences
(e.g. long gene sequences with more then 10 thousand characters that
are treated as one sentence). Base on the distribution of sentence
length (see Figure \ref{fig:The-distribution-of-sent-len}), we remove
the sentences that are outliers either in character or word length.
We winsorize 5\% and 95\% quantiles. For character-wise length, this
amounts to 19 characters for 5\% quantile and 275 characters for 95\%
quantile. For word-wise length, it is 3 words and 42 words, respectively.
\item \textbf{Hierarchical tree-like structure}. By using section and paragraph
tagging information in the XML file and the sentences we extracted
in previous step, we construct a hierarchical tree-like structure
of the articles. In this structure, sentences are contained within
paragraphs, which in turn are contained within sections. For each
section, we extract the section-type attribute from the XML file which
indicates which kind of section is (from a pre-defined set). For those
sections without a section-type, we use the section title instead.
\item \textbf{Citation hints removal.} The citing sentence usually has some
explicit hints which discloses a citation. This provides too much
information for the model training and it does not faithfully represents
a real-world application scenario. Thus, we removed all the citation
hints by regular expression (see Table \ref{tab:expression-to-remove-hints}).
\item \textbf{Noise removal. }There are other filters applied to remove
noise from sentences. We apply the following cleanup steps: trim white
spaces at beginning and end of a sentence, remove the numbers or punctuations
at the beginning of a sentence, and remove numbers at the end of a
sentence.
\end{APAenumerate}

\begin{table}
\caption{\label{tab:expression-to-remove-hints}regular expression to remove
the citation hints}

\centering{}%
\begin{tabular}{>{\centering}m{0.3\linewidth}>{\centering}m{0.2\linewidth}>{\centering}m{0.4\linewidth}}
\hline 
Regular expression & Description & Example\tabularnewline
\hline 
(?<!\textasciicircum )({[}\textbackslash{[}\textbackslash ({]}){[}\textbackslash s{]}{*}
({[}\textbackslash d{]}{[}\textbackslash s\textbackslash ,\textbackslash -\textbackslash --\textbackslash ;\textbackslash -{]}{*}){*}
{[}\textbackslash d{]}{[}\textbackslash s{]}{*}{[}\textbackslash{]}\textbackslash ){]} & numbers contained in parentheses and square brackets & ``{[}1, 2{]}'', ``{[} 1- 2{]}'', ``(1-3)'', ``(1,2,3)'', ``{[}1-3,
5{]}'', ``{[}8{]},{[}9{]},{[}12{]}'', ``( 1-2; 4-6; 8 )''\tabularnewline
{[}\textbackslash (\textbackslash{[}{]}\textbackslash s{*}({[}\textasciicircum\textbackslash (\textbackslash )\textbackslash{[}\textbackslash{]}{]}{*}
(((16|17|18|19|20) \textbackslash d\{2\}(?!\textbackslash d))| (et{[}\textbackslash . \textbackslash s\textbackslash\textbackslash xa0{]}{*}al\textbackslash .))
{[}\textasciicircum\textbackslash (\textbackslash ){]}{*})?{[}\textbackslash )\textbackslash{]}{]} & text within parentheses & ``(Kim and li, 2008)'', ``(Heijman , 2013b)'', ``(Tárraga , 2006;
Capella-Gutiérrez , 2009)'', ``(Kobayashi et al., 2005)'', ``(Richart
and Barron, 1969; Campion et al, 1986)'', ``(Nasiell et al, 1983,
1986)''\tabularnewline
et{[}\textbackslash . \textbackslash s\textbackslash\textbackslash xa0{]}+al{[}\textbackslash .\textbackslash s\textbackslash (\textbackslash{[}{]}{*}
((16|17|18|19|20)\textbackslash d \{2\}){*}{[})\textbackslash{]}
\textbackslash s{]}{*}(?=\textbackslash D) & remove et al. and the following years & ``et al.'', ``et al. 2008'', ``et al. (2008)''\tabularnewline
\hline 
\end{tabular}
\end{table}

After the processing, we get a dataset with approximately 309 million
sentences. However, due to the computational cost and in order to
make all of our analysis manageable, we randomly sample articles whose
sentences produce close to one million sentences. We further split
the one million sentences, 60\% for training, 20\% for validating,
and 20\% for testing. We present some characteristics of the whole
dataset and one million sentence sample in Table \ref{table:characteristics-pmoa}.

\begin{table}[ht]
\centering{}\caption{\label{table:characteristics-pmoa} Characteristics of the ACL-ARC
dataset, whole PMOA-CITE dataset and a sample of PMOA-CITE which contains
one million sentences }
\begin{tabular}{llll}
\hline 
Items & ACL-ARC & PMOA-CITE & PMOA-CITE sample\tabularnewline
\hline 
articles & N/A & 2,075,208 & 6,754\tabularnewline
sections & N/A & 9,903,173 & 32,198\tabularnewline
paragraphs & N/A & 62,351,079 & 202,047\tabularnewline
sentences & 1,228,052 & 309,407,532 & 1,008,042\tabularnewline
sentences without citations & 1142275 & 249,138,591 & 811,659\tabularnewline
sentences with citations & 85777 & 60,268,941 & 196,383\tabularnewline
average characters per sentence & 131 & 132 & 132\tabularnewline
average words per sentence & 22 & 20 & 20\tabularnewline
\hline 
\end{tabular}
\end{table}

\subsection{Text Representation}

Some of our models use different text representations predicting citation
worthiness. We now describe them.

\textbf{Bag of words (BoW) representation} This is a widely-used representation
where the order of words in the original text is ignored and only
word frequencies are maintained. While this representation is clearly
very simple (see \citet{harris1954distributional}), it has shown
remarkable performance in several tasks.

We follow the standard definition of term-frequency inverse term-frequency
(tf-idf) to construct our bag of words (BoW) representation \citep{manning2008introduction}.
Our BoW representation for a sentence $S$ which consists of $n$
words will therefore be the vector of all tf-idf values in document
$D_{i}$. 
\begin{equation}
\text{BoW}(S)=[\text{tf-idf}_{w_{1},D_{i}},...,\text{tf-idf}_{w_{n},D_{i}\cdot}]\label{eq:bow}
\end{equation}

Sometimes, it is not possible to capture some language subtleties
using singular words as tokens. For example, ``play football'' has
a different meaning than ``football play''. Therefore, it is common
to also keep track of combinations of words in what are known as $n$-gram
language models \citep{jurafsky2014speech}. In our analysis, we use
unigrams and bigrams to construct the bag-of-words representation.

\textbf{Topic modeling based (TM) representation} Topic modeling is
a machine learning technique whose goal is to represent a document
as a mixture of a small number of ``topics''. This reduces the dimensionality
needed to represent a document compared to bag-of-words. There are
several topic models available including Latent Semantic Analysis
(LSA) and Non-negative Matrix Factorization (NMF). In this paper,
we use Latent Dirichlet Allocation (LDA), which is one of the most
popular and well-motivated approaches (for discussions of its advantage,
see \citet{blei2003latent,blei2012probabilistic}; however, also see
some shortcomings in \citet{lancichinetti2015high}).

\textbf{Distributed word representation} While topic models can extract
statistical structure across documents, they do a relatively poor
job at extracting information within documents. In particular, topic
models are not meant to find contextual relationships between words.
Word embedding methods, in contrast, are based on the distributional
hypothesis which states that words that occur in the same context
are likely to have similar meaning \citep{harris1954distributional}.
The famous statement ``you shall know a word by the company it keeps''
by \citet{firth1957synopsis} is a concise guideline for word embedding:
a word could be represented by means of the words surrounding it.
In word embedding, words are represented as fixed-length vectors that
attempt to approximate their semantic meaning within a document.

There are several distributed word representation methods but one
of the most successful and well-known is GloVe by \citet{pennington2014glove}.
We use GloVe word vectors with 300 dimensions, pre-trained on 6 billion
tokens.

\subsection{Sentence features and contextual features}

After sentence processing, we can get features from the sentence itself
and its context. For sentence, we get text representation, the character-wise
sentence length, the word-wise sentence length, whether the previous
and next sentences have citations. We can also include contextual
features: section text, the features describing the previous sentence
and the next sentence, the cosine similarity between the current sentence
and the surrounding sentence. We normalized the features using maximum
absolute scaling for sparse features and standardization for dense
features before feeding them into the models. These sentence features
and contextual features should capture a large portion of the attributes
associated with citation location while keeping a high level of interpretability. 

\subsection{Evaluation metrics}

We use precision, recall and $F_{1}$ as metrics to evaluate the performance
of our models. They are defined as follows:

\begin{equation}
\mathrm{Precision}=\frac{\mathrm{tp}}{tp+fp}
\end{equation}

\begin{equation}
\mathrm{Recall}=\frac{\mathrm{tp}}{tp+fn}
\end{equation}

\begin{equation}
F_{1}=2\cdot\frac{\mathrm{Precision}\cdot\mathrm{Recall}}{\mathrm{Precision}+\mathrm{Recall}}
\end{equation}

where, the $tp$ denotes the number of the sentences predicted to
be citing sentence and they are indeed have citation; the $fp$ refer
to those predicted to be citing sentence but they don't have citation;
the $fn$ represents the number of sentences have citation but predicted
don't have citation; $Precision$, $Recall$ and $F_{1}$ varies from
0 (worst) to 1 (best). 

All the evaluation results reported in this paper are measured for
the minority class (citing sentences) label.

\section{An Attention-based BiLSTM architecture for citation worthiness}

In this section, we describe our new architecture for improving upon
the performance of classic statistical learning models presented above.
Importantly, these models might neglect some of the interpretability
but might pay large performance dividends. Generally, they do not
need hand-crafted features. At a high level, the architecture we propose
has the following layers (also Fig. \ref{fig:Network-architecture}):
\begin{APAenumerate}
\item Character embedding layer: encode every character in a word using
a bidirectional LSTM, and get a vector representation of a word.
\item Word embedding layer: convert the tokens into vectors by using pre-trained
vectors.
\item Encoder layer: use a bidirectional LSTM which captures both the forward
and backward information flow.
\item Attention layer: make use of an attention mechanism to interpolate
the hidden states of the encoder (explained below)
\item Contextual Features layer: obtain the contextual features by combining
features of section, previous sentence, current sentence, and next
sentence.
\item Classifier layer: use a multilayer perceptron to produce the final
prediction of citation worthiness.
\end{APAenumerate}
\begin{center}
\begin{figure}
\begin{centering}
\includegraphics[width=0.85\linewidth]{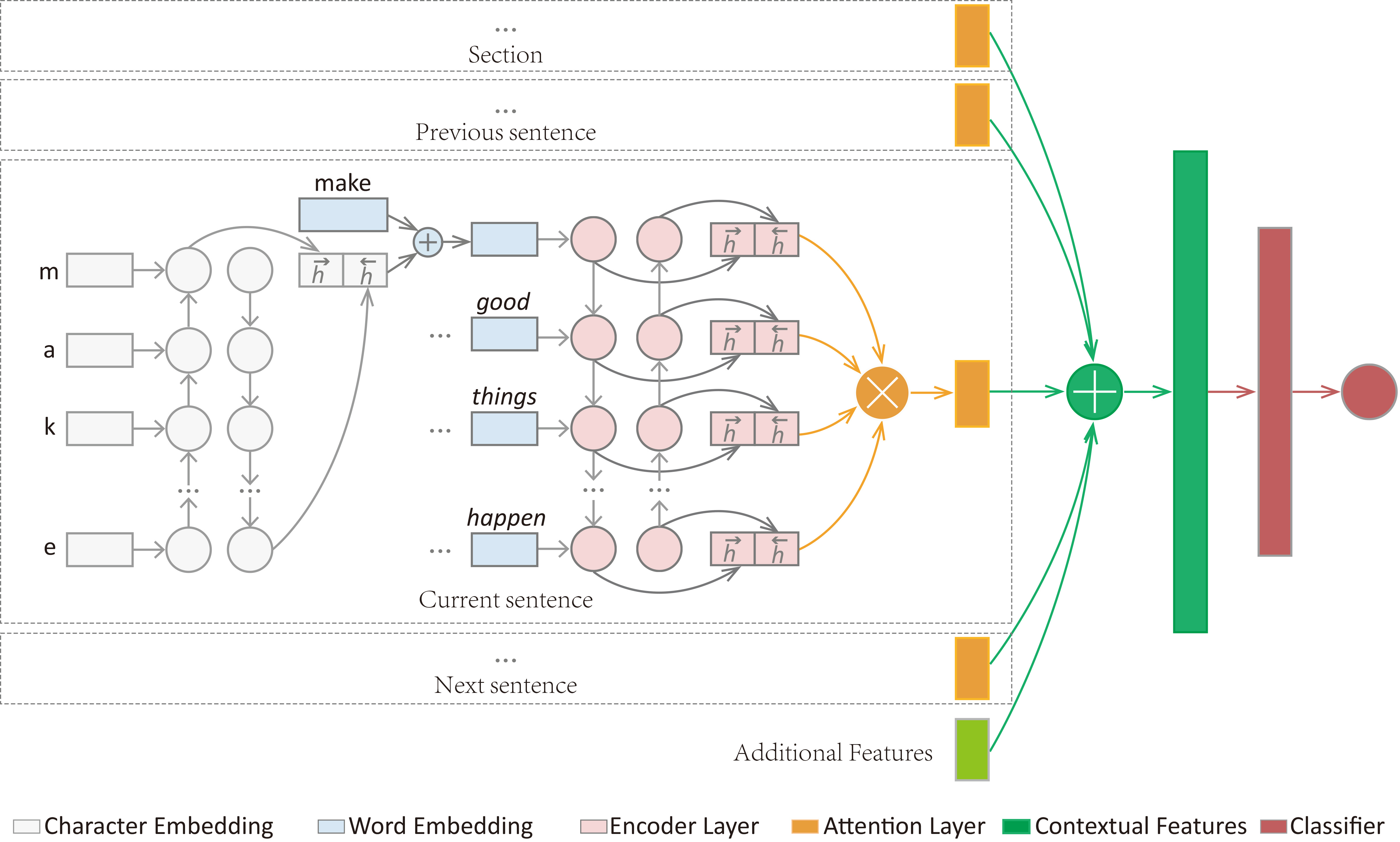}
\par\end{centering}
\caption{\label{fig:Network-architecture}The architecture of the proposed
attention-based BiLSTM neural network.}
\end{figure}
\par\end{center}

\subsection{Character Embedding}

In language modeling, it is a common practice to treat a word as the
basic unit. Similar to the bag of words assumption, we can consider
the text as composed of a bag of characters. \citet{santos2014learning},
\citet{zhangCharacterlevelConvolutionalNetworks2015} and \citet{chen2015joint}
shows that learning a character level embedding could benefit various
NLP tasks. In this layer, we get the characters from tokens, then
feed the characters to a bidirectional LSTM network, to get a fixed-length
representation of the token. By using the character embedding, we
can solve the out-of-vocabulary problem for pre-trained word embedding%
.

\subsection{Word Embedding}

The word embedding is responsible for mapping a word into a vector
of numbers which will be the input for the next layers. For a given
sentence $S$, we first convert it into a sequence consisting of $n$
tokens, $S=\{c_{1},c_{2},\cdots,c_{n},\}$ . For each token $c_{i}$,
we look up the embedding vector $x_{i}$ from a word embedding matrix
$M^{tkn}\in\mathbb{R}^{d|V|}$, where the $d$ is the dimension of
the embedding vector and the $V$ is the vocabulary size of the tokens.
In this paper, the matrix $M^{tkn}$ is initialized by pre-trained
GloVe vectors \citep{pennington2014glove}, but will be updated by
learning from our corpus. Before feeding the encoder, we concatenate
the word vectors from word embedding and character embedding.

\subsection{Encoder}

Recurrent neural networks (RNNs) are a powerful model to capture features
from sequential data, such as temporal series, and text. RNNs could
capture long-distance dependency in theory but they suffer from the
exploding/vanishing gradient problems \citep{pascanu2013difficulty}.
This is, as the network is unraveled, the training process becomes
chaotic. The Long short-term memory (LSTM) architecture was proposed
by \citet{hochreiter1997long} to solve these issues. LSTM introduces
several gates to control the proportion of information to forget from
previous time steps and to pass to the next time step. Formally, LSTM
could be described by the following equations:

\begin{equation}
i_{t}=\sigma(W_{i}x_{t}+W_{i}h_{t-1}+b_{i})
\end{equation}
\begin{equation}
f_{t}=\sigma(W_{f}x_{t}+W_{f}h_{t-1}+b_{f})
\end{equation}
\begin{equation}
g_{t}=tanh(W_{g}x_{t}+W_{g}h_{t-1}+b_{g})
\end{equation}
\begin{equation}
o_{t}=\sigma(W_{o}x_{t}+W_{o}h_{t-1}+b_{o})
\end{equation}
\begin{equation}
c_{t}=f_{t}\bigotimes c_{t-1}+i_{t}\bigotimes g_{t}
\end{equation}
\begin{equation}
h_{t}=o_{t}\bigotimes tanh(c_{t})
\end{equation}
where the $\sigma$ is the sigmoid function, $\bigotimes$ denotes
the dot product, $b$ is the bias, $W$ is the parameters, $x_{t}$
is the input at time $t$, $c_{t}$ is the LSTM cell state at time
$t$ and $h_{t}$ is hidden state at time $t$. The $i_{t}$, $f_{t}$,
$o_{t}$ and $g_{t}$ are called input, forget, output and cell gates
respectively, they control the information to keep in its state and
pass to next step.

LSTM gets information from the previous step, which is the context
to the left of the current token. However, it is important to consider
the information to the right of the current token. A solution of this
information need is bidirectional LSTM \citep{graves2013speech}.
The idea of BiLSTM is to use two LSTM layers and feed in each layer
with forward and backward sequences separately, concatenating the
hidden states of the two LSTM to model both contexts:

\begin{equation}
h_{t}=[\overrightarrow{h_{t}}\varoplus\overleftarrow{h_{t}}]
\end{equation}

For a sentence with $n$ tokens, the hidden state of BiLSTM $H$ would
be:

\begin{equation}
H=(h_{1},h_{2},...h_{n})\label{eq:hiddenstates}
\end{equation}

\subsection{Attention}

The \emph{attention} mechanism was introduced for sequence-to-sequence
models by \citet{bahdanauNeuralMachineTranslation2014}. They replace
the fixed context vector (produced by the encoder) with a dynamic
context vector which is a weighted average of the hidden state of
the encoder. The weight of each hidden state is determined by a score
between the encoder hidden states and the previous decoder states.

We can consider the previous decoder hidden state as a query vector,
the encoder hidden states as key and value vectors. In general, given
a query and set of key-value pairs, attention could be interpreted
as mapping the query to an output by using the weighted average of
the values. The weight to a certain value is computed by a score function
of the query with the corresponding key.

Formally, we denote the query as $q$, key as $(k_{1},k_{2},...,k_{n})$
and values as $(v_{1},v_{2},...,v_{n})$, the weigh vector as $(\alpha_{1},\alpha_{2},...,\alpha_{n})$,
both of them have same size $n$. the output $z$ is

\begin{equation}
z=\stackrel[i=1]{n}{\sum}\alpha_{i}v_{i}
\end{equation}

\begin{equation}
\alpha_{i}=\frac{\exp(\text{score}(q,k_{i}))}{\sum_{i^{'}=1}^{n}\exp(\text{score}(q,k_{i^{'}}))}
\end{equation}

The weight $\alpha_{i}$ is obtained by using the softmax function,
the $\text{score(\ensuremath{\cdot)}}$ is a compatibility function
between $q$ and $k_{i}$.

There are several score functions, such as additive \citep{bahdanauNeuralMachineTranslation2014}
and MLP \citep{linStructuredSelfattentiveSentence2017}, however,
these methods introduce more parameters to learn. In this paper, we
use the cosine ($cos$) score function introduced by \citet{graves2014neural},
dot-product ($dp$) score function introduced by \citet{luongEffectiveApproachesAttentionbased2015}
and the scaled dot-product ($sdp$) score function proposed by \citet{vaswaniAttentionAllYou2017a}.
These three approaches are a multiplication of two matrices with no
additional hyper-parameters:

\begin{equation}
score_{cos}(q,k)=\frac{q\cdot k}{||q||\text{·}||k||},\label{eq:cos-att}
\end{equation}

\begin{equation}
score_{dp}(q,k)=qk^{T},\label{eq:dp-att}
\end{equation}

\begin{equation}
score_{sdp}(q,k)=\frac{qk^{T}}{\sqrt{d_{k}}},\label{eq:sdp-att}
\end{equation}
where $d_{k}$ is the size of query vector. If there is a scale item,
a scalar $\sqrt{d_{k}}$ is applied, otherwise, the dot-product function
is applied.

In this research, the query is the hidden state of BiLSTM at the last
time step $H$ (see Eq.\ref{eq:hiddenstates}) of the encoder. By
using the attention mechanism, we effectively use all hidden states,
recovering long-distance information dependencies.

\subsection{Contextual Features}

In this layer, we concatenate the attention output of section, previous
sentence, current sentence and next sentence with 8 additional features.
The additional features including the character-wise and word-wise
sentence length for previous sentence, current sentence and next sentence
respectively, and whether previous and next sentence have citation.

\subsection{Classifying}

The last layer of our model is a classifier layer. The output of attention
layer $z$ is passed to a multilayer perceptron and then the softmax
function is applied to predict the probability of each class label
$\hat{y}$ for a given sentence $S$

\begin{equation}
p(y|S)=\text{softmax}(Wz+b)
\end{equation}

\begin{equation}
\hat{y}=\arg\max_{y}\hat{p}(y|S)
\end{equation}

We use the cross-entropy loss and L2 regularization as our cost function
to maximize the probability of true class label $\hat{y}$:
\begin{equation}
J(\theta)=-\stackrel[i]{N}{\sum}\stackrel[j]{C}{\sum}y_{j}^{(i)}\log\hat{y}_{j}^{(i)}+\lambda||\theta||^{2}
\end{equation}

Where $N$ is the total number of the training instances in a batch,
$C$ is the number of classes. $y$ is the ground-truth label indicator,
$\hat{y}$ is the probability of prediction. $\lambda$ is the amount
of L2 regularization, and $\theta$ represent all the trainable parameters.

\subsection{Network and training parameters}

For all Att-BiLSTM models, we set the dimension of hidden state for
character and word embedding to 15 and 128, respectively. We use RELU
as the activation function and Adam as the optimizer. We set the learning
rate to 0.001 and batch size to 64. In order to avoid over-fitting,
we set the dropout rate to 0.5, and also we use L2 regularization
of $10^{-7}$. During the training process, if the validation performance
does not improve for three epochs, we stop the training and choose
the model with best validation performance as the final model.

\section{Interpretable models for citation worthiness}

In this section, we want to introduce models which offer interpretable
results. 

\textbf{Elastic-net Regularized Logistic Regression}

The logistic regression model is as follows:

\begin{equation}
p(y_{i}=1)=\sigma(\beta^{T}\mathbf{x}+b),\label{eq:logistic-regression}
\end{equation}
where $\sigma(z)=(1+e^{-z})^{-1}$, $\beta$ are the weights, $x$
are the inputs, and $b$ is the intercept. If we use normalized terms
as features (e.g., tf-idf of uni-grams and bi-grams), we can directly
interpret the weights in Eq. \ref{eq:logistic-regression} to determine
whether a term increases the probably of citation or not, and by how
much. 

Most of the words in our dataset are not predictive of whether a citation
is needed. Therefore, we need to reduce the importance of them or
remove them from the prediction altogether. Elastic-net regularized
logistic regression (ENLR) allows to automatically perform both goals
\citep{friedman2001elements}. The ENLR loss function has the following
form
\begin{equation}
-\left[\sum_{i=1}^{N}y_{i}\cdot\log\sigma(\mathbf{x}_{i})+(1-y_{i})\cdot\log(1-\sigma(\mathbf{x}_{i}))\right]+\lambda\left[\frac{1}{2}\left(1-\alpha\right)||\beta||_{2}^{2}+\alpha||\beta||_{1}\right],\label{eq:enetlr}
\end{equation}
where $\sigma$ is the sigmoid function, $||\beta||_{2}^{2}$ is L2
regularization, and $||\beta||_{1}$ is L1 regularization. The parameter
$\alpha$, between 0 and 1, controls how much L1 regularization is
added and $\lambda$ controls how much of both regularizers are added
to the loss. The parameters $\alpha$ and $\lambda$ are chosen by
cross validation.

\textbf{Random Forest}

Elastic-net logistic regression can only find linear relationships
between the features $x$ and the output $y$. Random forest is a
general method for finding non-linear relationships by using bagging
of decision trees.  The final decision is made by averaging 
\begin{equation}
p(y_{i}=1)=\frac{1}{B}\sum_{j=1}^{B}T_{j}(\mathbf{x}_{i}),
\end{equation}
where $T_{j}$ is a decision tree fitted to a subset of the features
on a bootstrapped sample of the data. In the case of classification,
the final decision is made by majority vote. There are two parameters:
the number of trees, $B$, and the number of features to sample for
each tree, $p$. Both parameters are chosen by cross validation but
typically $p=\sqrt{m}$ where $m$ is the total number of features.
Empirical evidence suggests that random forest parameters are robust
to overfitting \citep{friedman2001elements}.

\section{Results}

In this work, we examined the factors that lead to a citation being
made. We propose a new dataset to answer this question and we proposed
a new method based on neural networks to predict which sentences need
a citation. We compare this method with several other techniques,
and interpret the findings.

\subsection{Attention-based BiLSTM model results}

We now examine high-performant models that are perhaps less interpretable.
We name models using features extracted from the current sentence
only as Att-BiLSTM with a subscript. We name models using the features
extracted from current sentence and its context as Contextual-Att-BiLSTM
with a subscript. The symbol $cos$, $dp$ and $sdp$ represent cosine
(Eq. \ref{eq:cos-att}), dot-product (Eq. \ref{eq:dp-att}) and scaled
dot-product (Eq. \ref{eq:sdp-att}) as the attention score function,
respectively.

\subsubsection{Results for ACL-ARC dataset}

\begin{table}
\centering{}\caption{\label{tab:Result-of-predicting}Comparison of our Att-BiLSTM models
with \citet{farber2018cite} and \citet{bonabCitationWorthinessSentences2018}.
The hyper-parameters of our models are chosen on the validation set
and the performances reported are base on a hold-out testing set.}
\begin{tabular}{cccc}
\hline 
Model & Precision & Recall & $F_{1}$\tabularnewline
\hline 
CNN GloVe & 0.196 & 0.269 & 0.227\tabularnewline
RNN GloVe & 0.171 & 0.317 & 0.222\tabularnewline
CRNN GloVe & 0.182 & 0.260 & 0.214\tabularnewline
CNN-rnd-update & 0.418 & 0.409 & 0.413\tabularnewline
CNN-w2v-update & 0.449 & 0.406 & 0.426\tabularnewline
\hline 
Att-BiLSTM$_{sdp}$ & 0.766 & 0.340 & 0.471\tabularnewline
Att-BiLSTM$_{dp}$ & 0.711 & 0.380 & 0.495\tabularnewline
Att-BiLSTM$_{cos}$ & 0.720 & 0.391 & \textbf{0.507}\tabularnewline
\hline 
\end{tabular}
\end{table}

\begin{figure}
\begin{centering}
\includegraphics[width=0.6\linewidth]{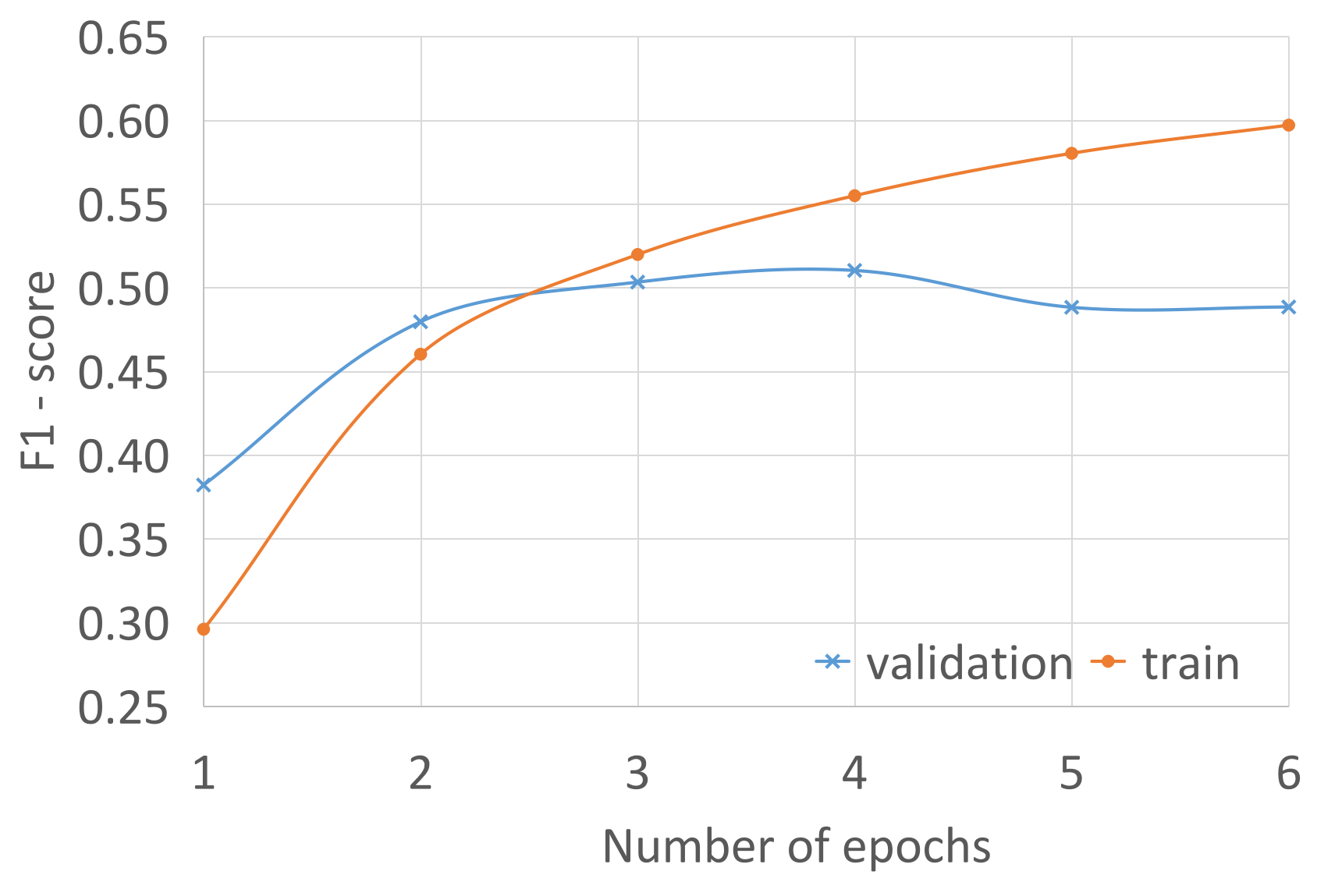}
\par\end{centering}
\caption{\label{fig:learning_curve_acl}The train and validation $F_{1}$ performance
for Att-BiLSTM$_{cos}$ using ACL-ARC dataset: x-axis shows the number
of epoch, the y-axis is $F_{1}$ score.}
\end{figure}

In this section, we first compare our models with the following state-of-art
approaches on this task as baselines:
\begin{APAitemize}
\item \textbf{CNN GloVe}:\textbf{ }a convolutional neural network with GloVe
word vector \citep{farber2018cite}.
\item \textbf{RNN GloVe}:\textbf{ }a recurrent neural network with GloVe
word vector \citep{farber2018cite}.
\item \textbf{CRNN GloVe}:\textbf{ }a convolutional recurrent neural network
approach proposed by \citep{farber2018cite}, using GloVe word vector.
\item \textbf{CNN-rnd-update}:\textbf{ }A CNN-based architecture proposed
by \citet{bonabCitationWorthinessSentences2018}, word embedding are
initialized randomly and updated during the training process.
\item \textbf{CNN-w2v-update}:\textbf{ }A CNN-based architecture proposed
by \citet{bonabCitationWorthinessSentences2018}, word embedding are
initialized by pre-trained word vectors and updated during the training
process.
\end{APAitemize}
Both of our models and the baselines are evaluated on the same ACL-ARC
corpus. In terms of recall, our approaches performs better than models
proposed by \citet{farber2018cite} but lower than models proposed
by \citet{bonabCitationWorthinessSentences2018}. However, in terms
of precision, our performance is much better than the baselines. Overall,
our results show that our model has significantly higher performance
than previous approaches (19\% more $F_{1}$). For our models, the
cosine score function has notably better performance against the dot-product
and scaled dot-product score function, but all of them are better
than baselines. As it is revealed by Figure \ref{fig:learning_curve_acl},
the neural network has find good scores relatively quickly.

\subsubsection{Results on PubMed Open Access subset dataset (PMOA-CITE)}

\begin{table}
\caption{\label{tab:result-neural-netowrk-pmoa}The best performance of models
on the PMOAS dataset. The hyper-parameters of our models are chosen
on the validation set and the performances reported are based on a
hold-out testing set.}

\centering{}%
\begin{tabular}{cccc}
\hline 
Model & Precision & Recall & $F_{1}$\tabularnewline
\hline 
Att-BiLSTM$_{sdp}$ & 0.900 & 0.764 & 0.827\tabularnewline
Att-BiLSTM$_{dp}$ & 0.886 & 0.788 & 0.834\tabularnewline
Att-BiLSTM$_{cos}$ & 0.883 & 0.795 & 0.837\tabularnewline
Contextual-Att-BiLSTM$_{sdp}$ & 0.907 & 0.797 & 0.848\tabularnewline
Contextual-Att-BiLSTM$_{dp}$ & 0.908 & 0.807 & 0.854\tabularnewline
Contextual-Att-BiLSTM$_{cos}$ & 0.907 & 0.811 & \textbf{0.856}\tabularnewline
\hline 
\end{tabular}
\end{table}

We would like to examine how our proposed architecture performs in
bio-medical fields with more contextual information added---the previous
ACL-ARC dataset does not contain contextual information. The results
are presented in Table \ref{tab:result-neural-netowrk-pmoa}. Similar
to before, the cosine attention score function performs best compared
with dot product and scaled dot product. This suggests that cosine
attention consistently outperforms the two attention score function
in different dataset, with or without contextual information. By adding
the contextual information, the testing performance improved significantly
(e.g. 0.837 to 0.856 in terms of $F_{1}$) across all the different
attention types. All our neural network models have much better performance
than statistical models discussed below (see Table \ref{table:lr_rf_perf}).

\begin{figure}
\centering{}\includegraphics[width=0.6\linewidth]{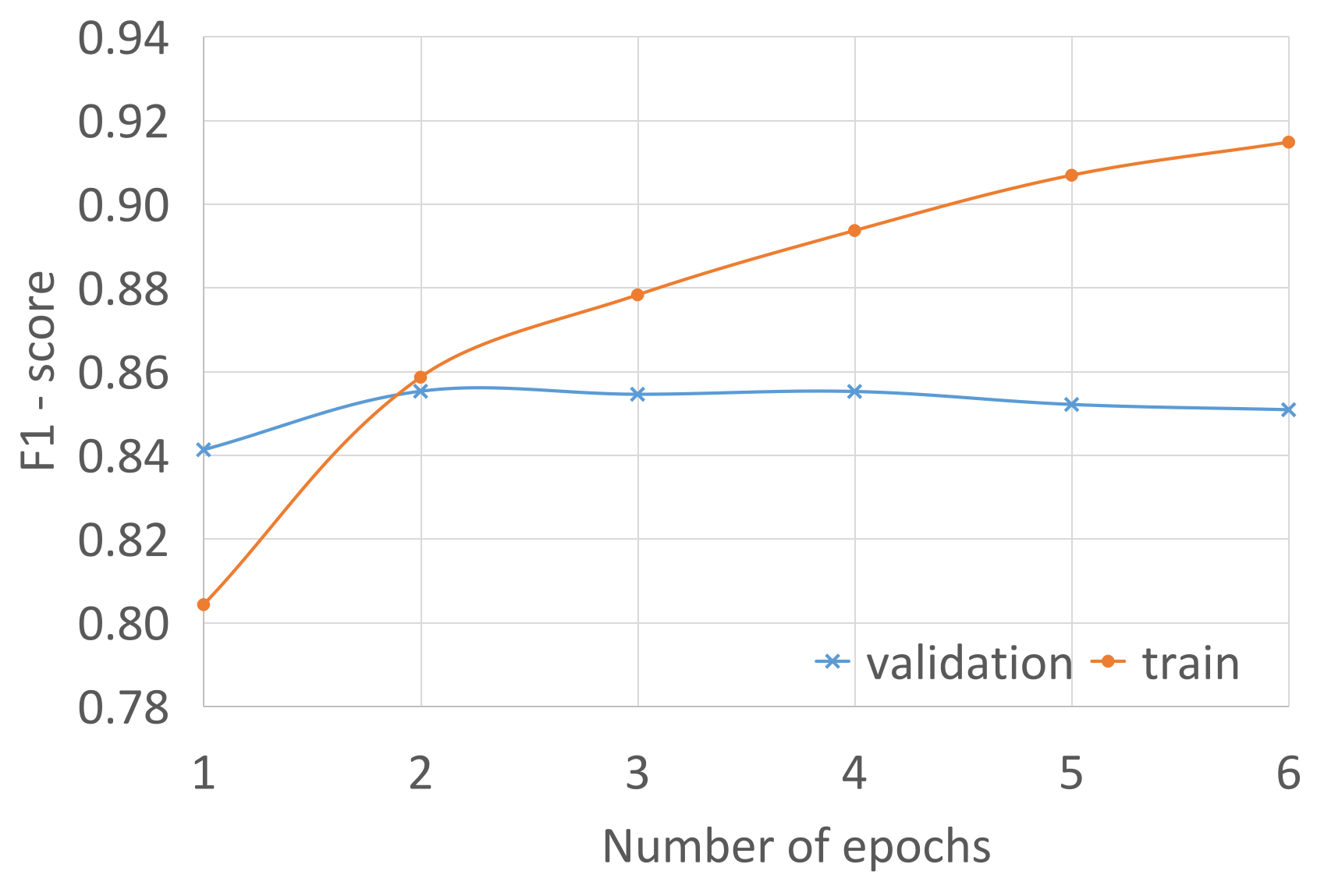}\caption{\label{fig:The-train-and}The train and validation $F_{1}$ performance
for Contextual-Att-BiLSTM$_{cos}$ using PMOAS dataset: x-axis shows
the number of epoch, the y-axis is $F_{1}$ score.}
\end{figure}

When we evaluate the training evolution of Contextual-Att-BiLSTM$_{cos}$
model (see Figure \ref{fig:The-train-and}), we found that the first
epoch gives enough information to achieve high performance. The validation
performance remains relatively stable after two epochs and reaches
its peak at the fourth epoch. Overall, our results suggest that deep
learning achieves considerable have better performance compared to
statistical models.

\subsubsection{Down-sampling sensitivity analysis}

\begin{figure}
\begin{centering}
\includegraphics[width=0.7\linewidth]{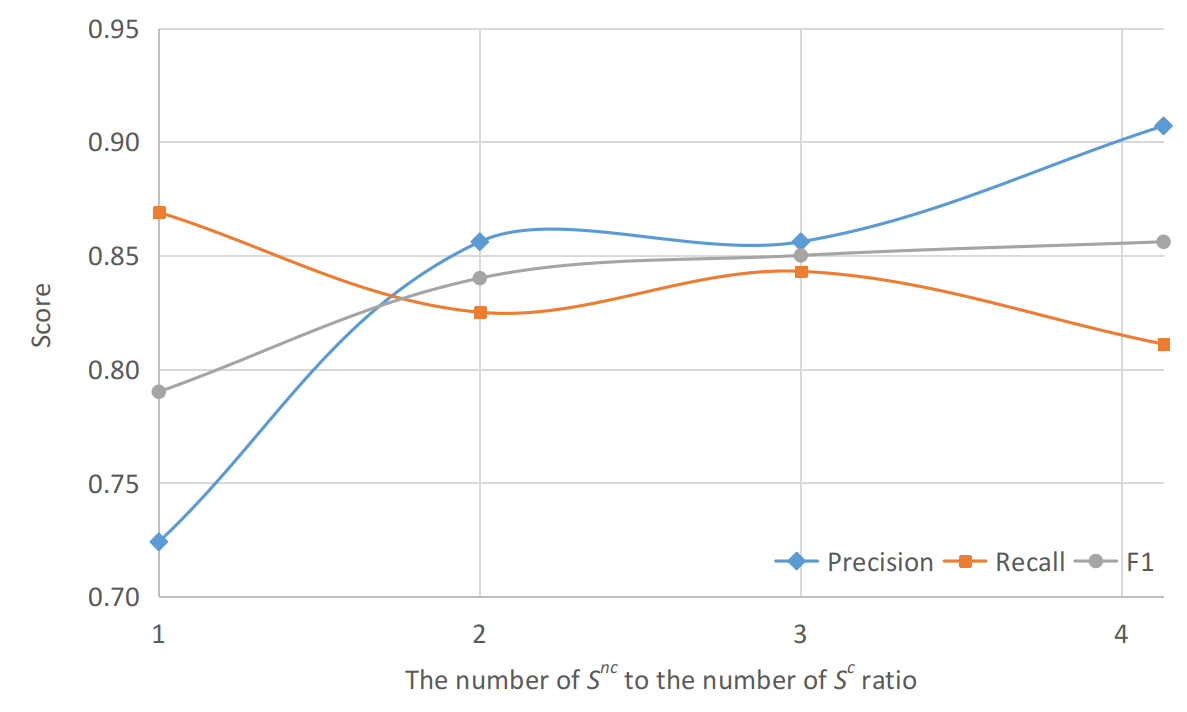}
\par\end{centering}
\caption{\label{fig:Down-sampling-sensitivity-analys}The performance of Att-BiLSTM$_{cos}$
when the PMOA-CITE dataset is down-sampled. $S^{nc}$ is the number
of citing sentences and $S^{c}$ is the number of citing sentences.}
\end{figure}

For most scientific papers, the number of citing sentence $S^{c}$
and non-citing sentences $S^{nc}$ are usually not equal and vary
across corpora. Therefore, this issue should be evaluated. In PMOA-CITE,
the number of $S^{nc}$ to the number of $S^{c}$ ratio is 4.13. A
typical approach to handle unbalanced dataset is to down-sample---reduce
the instances of the majority class and keep all the instance of minority
class. We investigate how the best architecture found in the previous
section (Att-BiLSTM$_{cos}$) would perform under these different
balance ratios. Importantly, the $S^{nc}$ to $S^{c}$ ratio of held-out
dataset remains the same as the natural proportion (4.13), following
standard information retrieval practice. Figure \ref{fig:Down-sampling-sensitivity-analys}
shows the performance under the ratios 1, 2, 3, and original 4.13.
The results suggest that our model has best performance when the ratio
equals the natural proportion. Also, increasing this ratio is associated
with an increase in the performance. Therefore, this sensitivity analysis
suggest that our model is robust to this kind of down-sampling sensitivity
analysis.

\subsubsection{Model generalization to different corpora}

Transfer learning has become an important topic in AI. Ideally, we
would like our model trained on PMOA-CITE, for example, to translate
to other datasets. Thus, we experimented with training on PMOA-CITE
and estimating performance on ACL-ARC, and vice versa. Expectedly,
generalization is hard as models trained and tested on the same dataset
have significantly better performance than training on one and evaluation
on the other (Table \ref{tab:transfer_perf}). However, the models
trained on PMOA-CITE have better cross-dataset generalization performance
than those trained on ACL-ARC. We attribute this improvement to the
overall quality of this dataset. The ACL-ARC dataset was extracted
from PDFs, which induces noise, while PMOA-CITE is already structured
with a pre-defined structure (e.g., tag set; see Materials and Methods).
Still, generalization is a challenging task.

We also perform experiments on training a model on a combination of
PMOA-CITE and ACL-ARC corpora. We matched the amount of data coming
from both sources by randomly sampling half from PMOA-CITE and half
from ACL-ARC training. Surprisingly, the testing performance on PMOA-CITE
and ACL-ARC are more than two times better than previously, and close
to the model training and testing solely on one dataset. This suggests
a way to solve the generalization problem by training the same model
with disparate domains.

\begin{table}
\caption{\label{tab:transfer_perf}The performance of model generalization}

\centering{}%
\begin{tabular}{ccccc}
\toprule 
Architecture & {\small{}Training $\rightarrow$ Testing corpora} & {\small{}Precision} & {\small{}Recall} & $F_{1}$\tabularnewline
\midrule
Att-BiLSTM$_{cosine}$ & ACL-ARC $\rightarrow$ ACL-ARC & 0.72 & 0.391 & 0.507\tabularnewline
Att-BiLSTM$_{cosine}$ & ACL-ARC $\rightarrow$ PMOA-CITE & 0.431 & 0.121 & 0.189\tabularnewline
Att-BiLSTM$_{cosine}$ & PMOA-CITE $\rightarrow$ PMOA-CITE & 0.883 & 0.795 & 0.837\tabularnewline
Att-BiLSTM$_{cosine}$ & PMOA-CITE $\rightarrow$ ACL-ARC & 0.155 & 0.289 & 0.202\tabularnewline
\midrule
Att-BiLSTM$_{cosine}$ & combined $\rightarrow$ ACL-ARC & 0.941  & 0.669 & 0.440\tabularnewline
Att-BiLSTM$_{cosine}$ & combined $\rightarrow$ PMOA-CITE & 0.860  & 0.765  & 0.809\tabularnewline
\bottomrule
\end{tabular}
\end{table}

\subsubsection{Real world applications of our algorithm for publishing}

We wanted to examine whether our model can find sentences that seem
to be misclassified but actually should have citations. We use the
model to make predictions for a hold-out testing set of 201,513 sentences,
3270 of them predicted to be cite worthy. We then manually examine
the top 10 sentences by cite worthiness probability (Table \ref{tab:some-mis-cite-examples-1}).
Surprisingly, the model discovered some mistakes that we suspect are
introduced by scientists or systems processing the accepted manuscripts.
These problems can be grouped into three categories. The first category
is an \textit{XML annotation error}. According to the JATS standard,
a bibliographic reference in a \emph{ref }tag should be denoted as
a \textit{bibr} property for \textit{ref-type} attribute. However,
cases numbered 1, 2, 7, and 9 in Table \ref{tab:some-mis-cite-examples-1}
do not comply with this standard and marked as non-citing sentence,
but our model detected these sentences should have a citation. A second
category contains \emph{citations not made properly}. In case numbered
8, the author puts an URL at the end of the sentence and therefore
the citation is not properly made. The third and perhaps more severe
category is \emph{mis-citations}. For the cases numbered 3, 4, 5 and
10, there are no cross reference annotations in the XML file. However,
based on the language, we think it would be better to cite the source
of the ideas. An interesting, borderline case outside of this categorization
is case 6, which could have a citation but it does not because the
authors cited the source somewhere else but felt that citing again
would be redundant. In sum, these manually verified cases show that
our model could indeed find mistakes. Furthermore, the probability
in the prediction could be used as an threshold for warning during
pre-submission or during peer review. Thus, our model could be used
as a filter for such mistakes before publishing.

\begin{table}
\caption{\label{tab:some-mis-cite-examples-1}some mis-citation examples}

\centering{}%
\begin{tabular}{c>{\centering}p{0.1\linewidth}>{\centering}p{0.3\linewidth}cc}
\hline 
Case number & Section & Sentence & Probability & Type\tabularnewline
\hline 
{\tiny{}1} & {\tiny{}introduction} & {\tiny{}The estimated cost of TBI in the United States is \$56 billion
annually , , with over 1.7 million people yearly suffering from TBI,
often resulting in undiagnosed pathology that can lead to chronic
disability , .} & {\tiny{}0.9999} & {\tiny{}XML annotation error}\tabularnewline
{\tiny{}2} & {\tiny{}discussion} & {\tiny{}It has been reported that the inferior turbinate and uncinate
process differ dramatically in levels of plasminogen activators and
host defense molecules , , .} & {\tiny{}0.9998} & {\tiny{}XML annotation error}\tabularnewline
{\tiny{}3} & {\tiny{}discussion} & {\tiny{}The importance of hsp90 for CpG ODN-PO--mediated signal transduction
is also suggested by Okuya , who recently showed that hsp90 converted
inert self-DNA or mainly CpG ODN-PO into potent triggers of IFN-\textgreek{a}
secretion .} & {\tiny{}0.9997} & {\tiny{}mis-citations}\tabularnewline
{\tiny{}4} & {\tiny{}discussion} & {\tiny{}The FZP gene and its orthologs in cereals participate in the
establishment of floral meristem identity, and fzp mutations affect
early events during spikelet development {[} ,} & {\tiny{}0.9996} & {\tiny{}mis-citations}\tabularnewline
{\tiny{}5} & {\tiny{}discussion} & {\tiny{}Most importantly, many of the reporter gene expression assays,
specially the one with GFP reporter gene15, may not differentiate
between the live and dead intracellular amastigotes.} & {\tiny{}0.9995} & {\tiny{}mis-citations}\tabularnewline
{\tiny{}6} & {\tiny{}intro} & {\tiny{}Importantly, Kessler and Rutherford found the strongest advantage
for visible over occluded responses at 60°, i.e. at the maximum overlap
between the avatar's and the egocentric LoS , reflecting an egocentric
influence on processing of the other's perspective.} & {\tiny{}0.9994} & {\tiny{}mentions after citation}\tabularnewline
{\tiny{}7} & {\tiny{}discussion} & {\tiny{}B. ovatus has previously been shown to utilize galactomannan
as a carbon source , .} & {\tiny{}0.9994} & {\tiny{}XML annotation error}\tabularnewline
{\tiny{}8} & {\tiny{}introduction} & {\tiny{}More than 600 cry genes have been described (http://www.lifesci.sussex.ac.uk/home/Neil\_Crickmore/Bt/toxins2.html).} & {\tiny{}0.9993} & {\tiny{}citations not made properly}\tabularnewline
{\tiny{}9} & {\tiny{}materials and methods} & {\tiny{}The bioinformatic pipeline used to extract TCR\textgreek{b}
sequences was described previously , .} & {\tiny{}0.9992} & {\tiny{}XML annotation error}\tabularnewline
{\tiny{}10} & {\tiny{}discussion} & {\tiny{}Consistent with previously stated studies, patients who underwent
angiography and embolization were reported to have significantly less
blood loss during surgery .} & {\tiny{}0.9992} & {\tiny{}mis-citations}\tabularnewline
\hline 
\end{tabular}
\end{table}

\subsection{Interpretable statistical model results}

Statistical learning usually comprises two main parts: prediction
and interpretation \citep{James:2014:ISL:2517747}. As it is well-known,
deep learning provides extremely good prediction performance by trading
it off with interpretation. We now explore more interpretable statistical
models that can help us understand the language promoting and inhibiting
citations worthiness.

\subsubsection{Comparison of models and features}

We perform experiments using Elastic-net regularized logistic regression
(ENLR) and Random Forest (RF) on the bag-of-words (BoW) representation
and topic modeling (TM) based representation, with and without contextual
information. We first examine the performance of ENLR. In our cross
validation, we sweep through a set of regularization parameters ($\lambda$)
and L1--L2 mixture parameters ($\alpha$, see Eq. \ref{eq:enetlr}).
Table \ref{table:lr_rf_perf} shows the test performance for the best
cross-validated models across these parameters for different features
sets. In general, the BoW representation achieves better overall performance
than TM representation; the contextual information improves the performance
significantly for both representations. The BoW representation with
contextual information gives the best performance ($F_{1}=0.581$).

\begin{table}[ht]
\begin{centering}
\caption{\label{table:lr_rf_perf}Elastic-net logistic regression (ENLR) model
performance}
\par\end{centering}
\centering{}%
\begin{tabular*}{1\linewidth}{@{\extracolsep{\fill}}c>{\centering}p{0.2\linewidth}>{\centering}p{0.25\linewidth}ccc}
\hline 
Model & Text representation & Feature set & Precision & Recall & $F_{1}$\tabularnewline
\hline 
$\text{ENLR}_{bow}$ & BoW & current sentence & 0.461 & 0.619  & 0.528\tabularnewline
$\text{ENLR}_{bowctx}$ & BoW & current sentence + contextual & 0.501 & 0.691 & \textbf{0.581}\tabularnewline
$\text{ENLR}_{tm}$ & TM & current sentence & 0.278  & 0.661  & 0.392\tabularnewline
$\text{ENLR}_{tmctx}$ & TM & current sentence + contextual & 0.378 & 0.624 & 0.471\tabularnewline
\hline 
$\text{RF}_{bow}$ & BoW & current sentence & 0.402 & 0.563 & 0.469\tabularnewline
$\text{RF}_{bowctx}$ & BoW & current sentence + contextual & 0.453 & 0.637 & \textbf{0.529}\tabularnewline
$\text{RF}_{tm}$ & TM & current sentence & 0.281 & 0.645 & 0.391\tabularnewline
$\text{RF}_{tmctx}$ & TM & current sentence + contextual & 0.398 & 0.594 & 0.477\tabularnewline
\hline 
\end{tabular*}
\end{table}

We then examine the performance of Random Forest, which has the ability
to capture non-linear relationships between features. We evaluate
RFs with 100, 200, and 500 trees with a parameter sampling strategy
that uses the square root of the number of features, $\sqrt{p}$.
A number of 500 trees was the best parameter for both BoW and topic
modeling. Table \ref{table:lr_rf_perf} shows the best performance
across these parameters on testing. The results suggest that, in contrast
to ENLR, RF performs best with BoW ($F_{1}=0.529$) representation
while TM representation has a lower performance ($F_{1}=0.477$).
As with ENLR, the topic models had the worst performance ($F_{1}=0.391$)
and the contextual information promotes the performance notably for
both representations.

\subsubsection{Word importance in the prediction}

Interpretation usually combines two steps: extraction of the important
features and the direction of influence of those features. For example,
we would like to know which words are most related to the presence
or absence of citations (e.g., feature importance) and which of these
words promotes (positive sign) or inhibits (negative sign) citation
worthiness. We get the feature importance from the random forest model
and the direction of the influence from an elastic net regularized
logistic regression model. Feature importance across all the features
sum up to one: the larger the feature importance, the more it affects
the model. Therefore, random forest and logistic regression can be
combined for the two steps necessary for interpretation.

First, we analyze the features at a high level by evaluating the combined
importance of terms. We sum the feature importance of all uni-grams
and bi-grams to form a category for the section type and current and
contextual sentences. In this manner, we know the overall influence
of these components before understanding the importance of the terms
they contain. This analysis shows that target sentence plays the most
import role (Table \ref{tab:feature-importance-for-bow}), followed
by next sentence and the section. The previous sentence has the smallest
impact. It is worth noting that the feature importance of current
sentence is more than 4.86, 4.94 4.30 times important than section
type, previous sentence and next sentence, respectively. However,
as the performance of the models showed above, still the context contributes
significant performance advantages. Also, we found that the more characters
or words a sentence has, the more likely the sentence needs a citation
(Table \ref{tab:feature-importance-for-bow} ). For individual features,
the presence of a citation in previous and next sentences has significant
positive impact on citation worthiness.

We wanted to understand how the section relates to citation worthiness.
Table \ref{table:regression-weights-section} shows the 10 most positive
and 10 most negative weights of the section type. The positive terms
in the section type are related to background information and discussion
(e.g. ``introduction'', ``background'', ``discussion'') where
scientific papers usually describe previous work to contextualize
the research being reported. In contrast, the negative terms are related
to descriptions and reports (e.g. ``results'', ``methods'', ``materials''),
therefore lowering the probability to have a citation. Thus, this
shows that the section can be have different influences on citation
worthiness.

We also wanted to investigate which uni-grams and bi-grams in the
current sentence relates to the citation worthiness. The positive
words are intuitive as they relate to describing events from the past
(e.g., \textquotedbl previously\textquotedbl , \textquotedbl previous\textquotedbl ,
\textquotedbl recently\textquotedbl ) and mentioning other studies
(e.g., \textquotedbl reported\textquotedbl , \textquotedbl described\textquotedbl ,
\textquotedbl been demonstrated\textquotedbl{} ). Negative terms
refer to the current paper (e.g., \textquotedbl this study\textquotedbl ,
\textquotedbl the study\textquotedbl ), entities which do not need
a citation (e.g., floating elements: \textquotedbl figure\textquotedbl ,
\textquotedbl table\textquotedbl ; statistics: \textquotedbl min\textquotedbl ,
\textquotedbl mean\textquotedbl , \textquotedbl test\textquotedbl ;
proper names: \textquotedbl usa\textquotedbl , \textquotedbl cells
were\textquotedbl ), and descriptive languages of experiments or
actions taken within the paper (e.g,. \textquotedbl washed\textquotedbl ,
\textquotedbl incubated\textquotedbl ). Therefore, uni-grams and
bi-grams and their weights reveal interesting patters about the presence
or absence of a citation.

\begin{table}
\caption{\label{tab:feature-importance-for-bow}Feature importance for BoW
representation. All the feature importance sum up to 1. The plus sign
(+) means the feature has a positive influence on the citation worthiness.
The minus sign (-) means the feature has a negative influence on the
citation worthiness.}

\centering{}%
\begin{tabular}{ccc}
\hline 
Category & Importance & Sign of Influence\tabularnewline
\hline 
sum of term importance of section type & 0.11627 & N/A\tabularnewline
sum of term importance of $S_{n-1}$  & 0.10967 & N/A\tabularnewline
number of characters in $S_{n-1}$  & 0.00279 & +\tabularnewline
number of words in $S_{n-1}$  & 0.00191 & +\tabularnewline
sum of term importance of $S_{n}$  & 0.54243 & N/A\tabularnewline
number of characters in $S_{n}$  & 0.01061 & +\tabularnewline
number of words in $S_{n}$  & 0.01230 & +\tabularnewline
sum of term importance of $S_{n+1}$  & 0.11840 & N/A\tabularnewline
number of characters in $S_{n+1}$  & 0.00803 & +\tabularnewline
number of words in $S_{n+1}$  & 0.00492 & +\tabularnewline
similarity between $S_{n-1}$ and $S_{n}$ & 0.00566 & -\tabularnewline
similarity between $S_{n+1}$ and $S_{n}$ & 0.00694 & -\tabularnewline
whether $S_{n-1}$ has citation & 0.02461 & +\tabularnewline
whether $S_{n+1}$ has citation & 0.03545 & +\tabularnewline
\hline 
\end{tabular}
\end{table}

\begin{table}
\begin{centering}
\caption{\label{table:regression-weights-section} Term (uni-gram or bi-gram)
importance of the section type. The plus sign (+) means the feature
has a positive influence on the citation worthiness. The minus sign
(-) means the feature has a negative influence on the citation worthiness.}
\par\end{centering}
\centering{}%
\begin{tabular}{ccc}
\hline 
Term in Section Type & Feature Importance & Sign of Influence\tabularnewline
\hline 
introduction & 0.027079 & +\tabularnewline
intro & 0.015247 & +\tabularnewline
background & 0.008828 & +\tabularnewline
discussion & 0.003698 & +\tabularnewline
cancer & 0.00155 & +\tabularnewline
mechanisms & 0.001328 & +\tabularnewline
cells & 0.000695 & +\tabularnewline
role & 0.000562 & +\tabularnewline
cell & 0.000521 & +\tabularnewline
receptors & 0.000379 & +\tabularnewline
\hline 
\end{tabular} %
\begin{tabular}{ccc}
\hline 
Term in Section Type & Feature Importance & Sign of Influence\tabularnewline
\hline 
results &  0.0174270  & -\tabularnewline
methods &  0.0128930  & -\tabularnewline
materials &  0.0059190  & -\tabularnewline
case &  0.0024290  & -\tabularnewline
report &  0.0014750  & -\tabularnewline
experimental &  0.0010170  & -\tabularnewline
authors &  0.0008160  & -\tabularnewline
conclusion &  0.0007460  & -\tabularnewline
presentation &  0.0006070  & -\tabularnewline
contributions &  0.0006050  & -\tabularnewline
\hline 
\end{tabular}
\end{table}

\begin{table}
\begin{centering}
\caption{\label{table:regression-weights-mid} Term importance of current sentence.
The plus sign (+) means the feature has a positive influence on the
citation worthiness. The minus sign (-) means the feature has a negative
influence on the citation worthiness.}
\par\end{centering}
\centering{}%
\begin{tabular}{ccc}
\hline 
Term in Section Type & Feature Importance & Sign of Influence\tabularnewline
\hline 
previously & 0.020892 & +\tabularnewline
has been & 0.015886 & +\tabularnewline
reported & 0.013642 & +\tabularnewline
previous & 0.012692 & +\tabularnewline
studies & 0.011627 & +\tabularnewline
been reported & 0.010001 & +\tabularnewline
shown that & 0.009809 & +\tabularnewline
have been & 0.009625 & +\tabularnewline
reported that & 0.009573 & +\tabularnewline
previously described & 0.009507 & +\tabularnewline
described & 0.009468 & +\tabularnewline
recently & 0.008796 & +\tabularnewline
recent & 0.008359 & +\tabularnewline
been shown & 0.007396 & +\tabularnewline
studies have & 0.006598 & +\tabularnewline
previous studies & 0.006528 & +\tabularnewline
described previously & 0.004926 & +\tabularnewline
associated with & 0.00485 & +\tabularnewline
been demonstrated & 0.004735 & +\tabularnewline
known & 0.004124 & +\tabularnewline
\hline 
\end{tabular} %
\begin{tabular}{ccc}
\hline 
Term in Section Type & Feature Importance & Sign of Influence\tabularnewline
\hline 
figure & 0.00224 & -\tabularnewline
table & 0.002148 & -\tabularnewline
fig & 0.001865 & -\tabularnewline
samples & 0.001642 & -\tabularnewline
participants & 0.001423 & -\tabularnewline
cells were & 0.001363 & -\tabularnewline
this study & 0.001358 & -\tabularnewline
pbs & 0.001152 & -\tabularnewline
min & 0.000894 & -\tabularnewline
the study & 0.000816 & -\tabularnewline
washed & 0.000812 & -\tabularnewline
mean & 0.00072 & -\tabularnewline
test & 0.00072 & -\tabularnewline
groups & 0.000699 & -\tabularnewline
difference & 0.000698 & -\tabularnewline
for min & 0.000673 & -\tabularnewline
sample & 0.00066 & -\tabularnewline
total & 0.000613 & -\tabularnewline
incubated & 0.000606 & -\tabularnewline
usa & 0.000573 & -\tabularnewline
\hline 
\end{tabular}
\end{table}

\subsubsection{Topic importance in the prediction}

In the section, we want to examine how the topic relates to citation
worthiness. Similar to above, we get the feature importance from random
forest model and the direction of feature influence from elastic net
regularized logistic regression model. We sum the feature importance
of all topics to form a category for the section type and current
and contextual sentences. As Table \ref{tab:feature-importance-for-topic-modeling}
shows, the current sentence has the largest impact on citation worthiness.
It is 2.06, 1.83 and 1.80 times more important than section type,
previous sentence and next sentence, respectively. Similar to the
BoW representation, the presence of citation in previous and next
sentence plays an import role as they are the top 2 most import features
across all the features.

\begin{table}
\caption{\label{tab:feature-importance-for-topic-modeling}Feature importance
for topic modeling representation. All the Feature Importance sum
to 1. he plus sign (+) means the feature has a positive influence
on the citation worthiness. The minus sign (-) means the feature has
a negative influence on the citation worthiness.}

\centering{}%
\begin{tabular}{ccc}
\hline 
Category & Importance & Sign of Influence\tabularnewline
\hline 
sum of topic importance of section type & 0.16323 & N/A\tabularnewline
sum of topic importance of $S_{n-1}$ & 0.18016 & N/A\tabularnewline
number of characters in $S_{n-1}$ & 0.00243 & +\tabularnewline
number of words in $S_{n-1}$ & 0.00185 & -\tabularnewline
sum of topic importance of $S_{n}$ & 0.31768 & N/A\tabularnewline
number of characters in $S_{n}$ & 0.00947 & -\tabularnewline
number of words in $S_{n}$ & 0.01049 & +\tabularnewline
sum of topic importance of $S_{n+1}$ & 0.18135 & N/A\tabularnewline
number of characters in $S_{n+1}$ & 0.00377 & +\tabularnewline
number of words in $S_{n+1}$ & 0.00223 & -\tabularnewline
similarity between $S_{n-1}$ and $S_{n}$ & 0.00272 & -\tabularnewline
similarity between $S_{n+1}$ and $S_{n}$ & 0.00327 & -\tabularnewline
whether $S_{n-1}$ has citation & 0.05891 & +\tabularnewline
whether $S_{n+1}$ has citation & 0.06244 & +\tabularnewline
\hline 
\end{tabular}
\end{table}

In order to understand a topic, we extracted all the terms and their
weights from the trained LDA model. The term weights for a topic are
a probability distribution and therefore sum up to 1. The large the
weight, the more important the term in the topic.

As Table \ref{tab:Topic-importance-of-section-type} shows, all the
topics in section type contribute positively to citation worthiness.
The most representative terms in each topic are methods, introduction,
conclusions and results. Across topics, there are some inflectional
forms of a word a word (e.g., ``materials'', ``material''). This
is because most section types are just one word, offering little information
to LDA.

\begin{table}
\caption{\label{tab:Topic-importance-of-section-type}Topic importance of the
section type }

\centering{}%
\begin{turn}{90}
\begin{tabular}{cccccccc}
\hline 
Topic Number & Importance & Sign & \multicolumn{5}{c}{Weights of Topic Terms}\tabularnewline
\hline 
\multirow{2}{*}{1} & \multirow{2}{*}{0.0588} & \multirow{2}{*}{+} & Terms & methods & materials & case & report\tabularnewline
\cline{4-8} \cline{5-8} \cline{6-8} \cline{7-8} \cline{8-8} 
 &  &  & Term Weights & 0.2155 & 0.1724 & 0.0199 & 0.0175\tabularnewline
\hline 
\multirow{2}{*}{2} & \multirow{2}{*}{0.0477} & \multirow{2}{*}{+} & Terms & introduction & background & authors & contributions\tabularnewline
\cline{4-8} \cline{5-8} \cline{6-8} \cline{7-8} \cline{8-8} 
 &  &  & Term Weights & 0.1239 & 0.0552 & 0.0114 & 0.0091\tabularnewline
\hline 
\multirow{2}{*}{0} & \multirow{2}{*}{0.0330} & \multirow{2}{*}{+} & Terms & conclusions & material & study & method\tabularnewline
\cline{4-8} \cline{5-8} \cline{6-8} \cline{7-8} \cline{8-8} 
 &  &  & Term Weights & 0.0633 & 0.0148 & 0.0114 & 0.0110\tabularnewline
\hline 
\multirow{2}{*}{3} & \multirow{2}{*}{0.0238} & \multirow{2}{*}{+} & Terms & results & discussion & intro & experimental\tabularnewline
\cline{4-8} \cline{5-8} \cline{6-8} \cline{7-8} \cline{8-8} 
 &  &  & Term Weights & 0.1998 & 0.1911 & 0.0714 & 0.0302\tabularnewline
\hline 
\end{tabular}
\end{turn}
\end{table}

We then further investigate the topic importance of the current sentence.
Table \ref{tab:Topic-importance-of-current-sentence} shows the three
most important positive and negative topics, with some arbitrary identifier.
The most important topic, topic 80 is represented by the terms describing
previous work. The second most important topic, topic 108 is represented
by bio-medical terms. Finally, topic 82 refers to methods and tools.
The importance of topic 80 doubles that of topics 108 and 82. This
suggests, similar to our BoW analysis above, that terms describing
previous work is highly related to citation worthiness. Conversely,
the most negative topics are 150, 122 and 179, and they all have similar
importance. These topics describe entities within the same paper,
(e.g., ``test'', ``fig'') which usually do not require citations.
Therefore, positive and negative topics have a great deal of interpretability.

\begin{table}
\caption{\label{tab:Topic-importance-of-current-sentence}Topic importance
of target sentence}

\centering{}%
\begin{turn}{90}
\begin{tabular}{cccccccc}
\hline 
Topic Number & Importance & Sign & \multicolumn{5}{c}{Topic Terms}\tabularnewline
\hline 
\multirow{2}{*}{80} & \multirow{2}{*}{0.0115} & \multirow{2}{*}{+} & Terms & described & previously & method & determined\tabularnewline
\cline{4-8} \cline{5-8} \cline{6-8} \cline{7-8} \cline{8-8} 
 &  &  & Term Weights & 0.0998 & 0.0830 & 0.0719 & 0.0640\tabularnewline
\hline 
\multirow{2}{*}{108} & \multirow{2}{*}{0.0055} & \multirow{2}{*}{+} & Terms & cells & cell & induced & following\tabularnewline
\cline{4-8} \cline{5-8} \cline{6-8} \cline{7-8} \cline{8-8} 
 &  &  & Term Weights & 0.1009 & 0.0680 & 0.0319 & 0.0296\tabularnewline
\hline 
\multirow{2}{*}{82} & \multirow{2}{*}{0.0052} & \multirow{2}{*}{+} & Terms & usa & analyses & version & spss\tabularnewline
\cline{4-8} \cline{5-8} \cline{6-8} \cline{7-8} \cline{8-8} 
 &  &  & Term Weights & 0.1331 & 0.0802 & 0.0717 & 0.0367\tabularnewline
\hline 
\multirow{2}{*}{150} & \multirow{2}{*}{0.0063} & \multirow{2}{*}{-} & Terms & test & fig & analyzed & post\tabularnewline
\cline{4-8} \cline{5-8} \cline{6-8} \cline{7-8} \cline{8-8} 
 &  &  & Term Weights & 0.0853 & 0.0690 & 0.0466 & 0.0428\tabularnewline
\hline 
\multirow{2}{*}{122} & \multirow{2}{*}{0.0043} & \multirow{2}{*}{-} & Terms & buffer & constant & nacl & contribution\tabularnewline
\cline{4-8} \cline{5-8} \cline{6-8} \cline{7-8} \cline{8-8} 
 &  &  & Term Weights & 0.0936 & 0.0544 & 0.0368 & 0.0361\tabularnewline
\hline 
\multirow{2}{*}{179} & \multirow{2}{*}{0.0036} & \multirow{2}{*}{-} & Terms & university & approved & minutes & committee\tabularnewline
\cline{4-8} \cline{5-8} \cline{6-8} \cline{7-8} \cline{8-8} 
 &  &  & Term Weights & 0.0658 & 0.0618 & 0.0514 & 0.0423\tabularnewline
\hline 
\end{tabular}
\end{turn}
\end{table}

\section{Discussion}

\begin{figure}[ht]
\begin{centering}
\includegraphics[width=12cm]{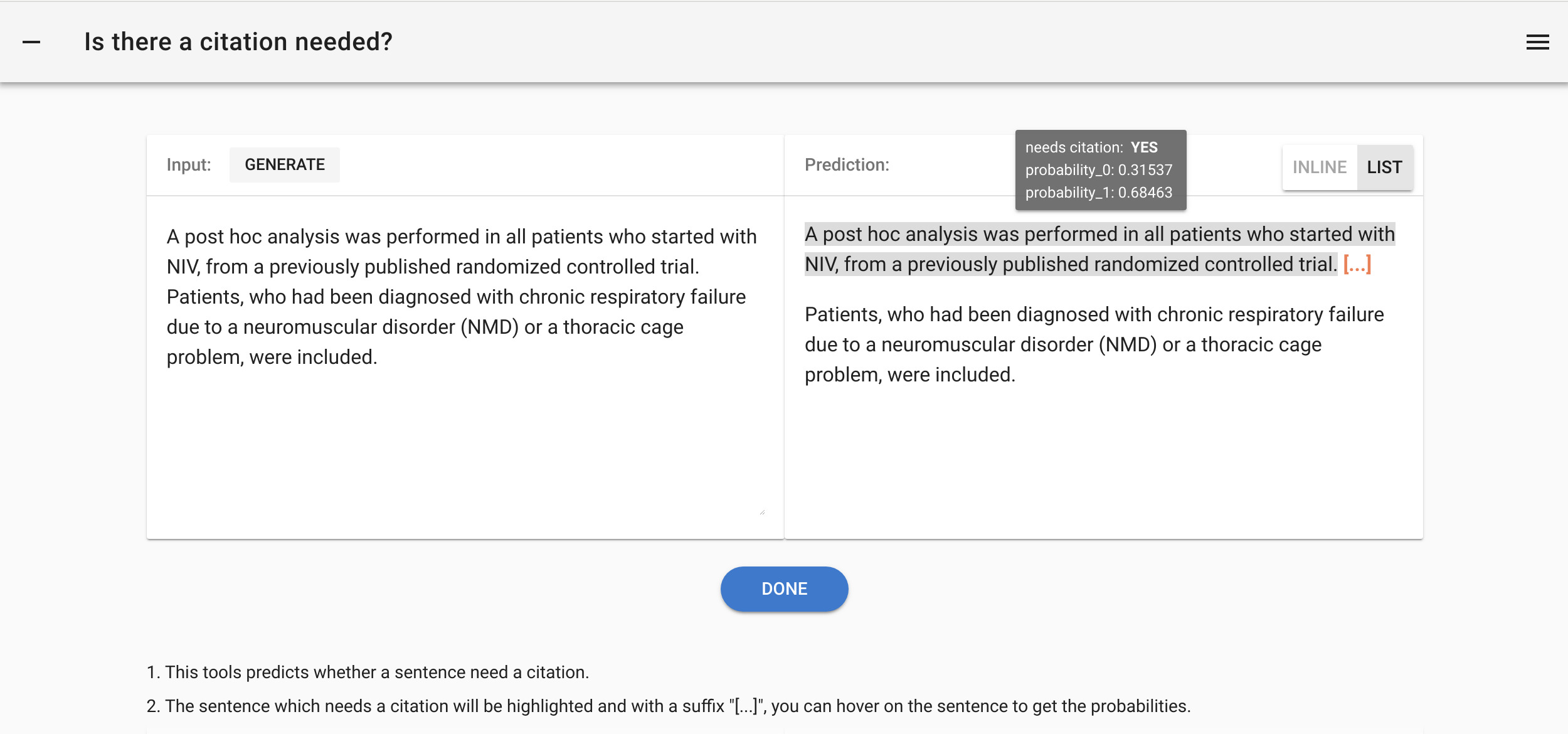}
\par\end{centering}
\caption{\label{fig:webtool}A screenshot of the online predicting tool}
\end{figure}

In this work, we developed methods and a large dataset for improving
the detection of citation worthiness. Citation worthiness is an important
first step for constructing robust and well-structured arguments in
science. It is crucial for determining where sources of ideas should
mentioned within a manuscript. Previous research has shown promising
results but thanks to our new large dataset and modern deep learning
architecture, we were able to achieve significantly good performance.
We additionally proposed several techniques to interpret what makes
scientists use citations. We uncovered potential issues in citation
data and behavior: XML documents not properly tagged, citations in
the wrong form, and, even worse, scientists failing to cite when they
should have. We make our code and a web-based tool available for the
scientific community. Our results and new datasets should contribute
to the larger need to complement scientific writing with automated
techniques. Taken together, our results suggest that deep learning
with modern attention-based mechanisms can be effectively used for
citation worthiness. We now describe contributions in the context
of other work and potential limitations of our approach.

The experimental results show that our proposed attention-based BiLSTM
architecture can effectively learn from the data. Compared with previous
state-of-art, our approach has significantly better performance ($F_{1}(\text{Att-BiLSTM}_{cos})=0.507$
vs $F_{1}(\text{CNN-w2v-update})=0.426$) in Table \ref{tab:Result-of-predicting}.
We attribute the improvements to: 1) The character embedding providing
extra information, 2) the effectiveness of the BiLSTM network to capture
sequential patterns in sentences, and 3) the attention mechanism helping
to generate better representations. When compared with interpretable
statistical models, our deep learning architecture has a large improvement
in $F_{1}$ (0.856 vs 0.581). This shows that the deep learning architecture
was better than the classical methods in terms of performance, but
at the cost of interpretability. Recent work by \citep{linStructuredSelfattentiveSentence2017,bahdanauNeuralMachineTranslation2014},
however, shows promising visualization techniques for attention mechanisms
that could improve deep learning interpretability for this task. Taken
together, our results suggest that deep learning with modern attention-based
mechanisms can be effectively used for citation worthiness.

As an enhancement to the ACL-ARC dataset, we proposed the PMOA-CITE
dataset in the hope of facilitating research on the citation worthiness
task. This extends the datasets available to the field of bio-medical
science. Our improvements are 1) a two orders of magnitude increase
in data size, 2) a well-structured XML file that is less noisy, and
3) contextual information. This dataset could be potentially used
in other citation context-related research, such as text summarization
\citep{chen2019automatic}, or citation recommendation \citep{HuangNeuralProbabilisticModela}.
Therefore, our contribution goes beyond the application of citation
worthiness.

Based on the experiments on PMOA-CITE dataset, the use of contextual
features consistently improved the performance. This improvement was
independent of the algorithm and text representation used (Tables
\ref{tab:result-neural-netowrk-pmoa} and \ref{table:lr_rf_perf})
. A similar results was reported in \citet{HeContextawareCitationRecommendation2010}
and \citet{jochim2012towards}). This suggests that contextual information
was key for citation worthiness and other related tasks.

The results of interpretable models reveal interesting patterns about
language use in citation worthiness. As suggested by writing guides
(e.g., \citet{booth2016craft}), researchers usually develop their
ideas based on several sources, which they quote, paraphrase, or summarize
and should therefore cite properly. Otherwise, the paper could lose
the trust of the readers, or even worse, run into suspicion of plagiarism
\citep{masic2013importance}. Our interpretation of feature importance
discovered some patterns of language usage for quoting, paraphrasing
or summarizing, and it real world applications could help to understanding
when a citation should be placed automatically. Also, the terms discovered
could help as educational resource for scientific writing. The interpretable
models then recognize systematic patterns of citations that are worth
exploring.

To the best of our knowledge, Table \ref{tab:transfer_perf} was the
first domain generalization performance reported for this task. While
the generalization was poor, this exercise highlights importance of
the domain knowledge to this task and showed the difficulty for domain
generalization. Our experiment showed that learning from multiple
source domains could promote the generalization on unseen target domains.
Thus, the release of our dataset can help in this endeavor.

In order to facilitate future research, we made our datasets and models
available to the public. The links of the dataset and the code parsing
XML files are available at \url{https://github.com/sciosci/cite-worthiness}
. We also built a web-based tool (see Figure \ref{fig:webtool}) at
\url{http://cite-worthiness.scienceofscience.org}. This tool might
help inform journalist, policy makers, the public to better understand
the principles of proper source citation and credit assignment.

We now discuss some of the limitations of our work. Our proposed deep
learning architecture is computational costly. We had to limit data
sample size used from our proposed dataset, and also we had to limit
the hyper-parameter search due to time constraints. These limitations
are mainly due to the RNN component of the architecture, whose encoding
and sequential nature were difficult to parallelize. One possible
solution to this problem could be the transformer network proposed
by \citet{vaswaniAttentionAllYou2017a}, because it eliminates recurrence
entirely, making it more parallelizable. However, it is unclear whether
the transformer could improve memory usage. Our validation performance
as a function of epochs, however, showed that the model was able to
learn relatively quickly, making it unclear whether more data would
significantly improve this already good performance (Fig. \ref{fig:The-train-and}).
In the future, we will investigate optimizations to our architecture
to improve its memory and time consumption.

When extracting the contextual features, we used previous sentence
and next sentence statically. However, there could be longer term
dependencies (e.g., information more than one sentence away) that,
when not included, incurred in contextual information loss. Conversely,
if the surrounding sentences truly were not semantically related to
the current sentence, adding them to the prediction could only produce
noise. As previously reported by \citet{kang2012characteristics},
only 5.2\% citations are multi-sentence. Although our approach has
some limitations, it still covers most situations and simplifies the
problem. To solve this, a possible solution could be identify the
sentences that are closely related to current sentence dynamically
\citep{kaplan2016citation,fetahufine,jebari2018new}. We will explore
this approach in the future.

Our approach was primarily developed with scientific articles from
bio-medical sciences in mind. Therefore, generalization to other domains,
such as news or general public pieces, can be severely limited. Scientific
writing might not be reflective of how journalists or other people
write other types of text. Science has a well-established set of rules
for adding citations, perhaps making the data ``too clean.'' Future
work will cross validate our results with general venues such as Wikipedia
(e.g., see \citet{chen2012citation}).

There are several avenues for future research. We can first investigate
how to recommend citations automatically based on the target sentence
and its context. This work offers the probability of a sentence need
a citation, thus forms an important step in this direction. In the
larger context of this research, there is a need to appropriately
cite credible sources. The research proposed here only addresses a
small portion of this challenge: while citation mistakes have been
estimated to be surprisingly prevalent---more than 20\% of citations
are wrong \citep{lukic2004citation,mogull2017accuracy}---we believe
that scientists tend to cite credible sources and they unintentionally
mis-cite. However, the problem is much more complex in other areas
such as \emph{fake news. }These types of articles \emph{do} cite sources
but the sources are not credible or taken out of context \citep{allcott2017social}.
Studies on detecting the credibility and quality of sources is a much
more complex problem which forms a challenging future research program.

\section{Conclusion}

In this article, we use open access scientific publications to detect
which sentences need citations. In particular, we build an deep learning
model based on attention mechanism and BiLSTM which achieve the state
of art performance while we also build models offering good interpretability.
We make the dataset, the model, and a web-based tool openly available
to the community. Our work therefore is an important step to improve
the quality of information and provide a data-driven tool to study
citations in science. We therefore hope that our work creates more
systematic studies regarding citation worthiness as it is the first
and crucial step for several tasks to make science more robust and
well-structured.

\section{Acknowledgments}

Tong Zeng was funded by the China Scholarship Council \#201706190067.
Daniel E. Acuna was partially funded by the National Science Foundation
awards \#1800956.

\bibliographystyle{apa}
\bibliography{sigproc2}

\end{document}